\documentclass[preprint,12pt]{elsarticle}

\usepackage{amsmath,amsfonts}
\usepackage{algorithmic}
\usepackage{array}
\usepackage[tight,footnotesize]{subfigure}
\usepackage{textcomp}
\usepackage{stfloats}
\usepackage{url}
\usepackage{verbatim}
\usepackage{graphicx}
\def\BibTeX{{\rm B\kern-.05em{\sc i\kern-.025em b}\kern-.08em
    T\kern-.1667em\lower.7ex\hbox{E}\kern-.125emX}}
\usepackage{balance}
\usepackage{booktabs}
\usepackage{caption}
\usepackage{makecell}
\usepackage{adjustbox}
\usepackage{wrapfig}
\usepackage[linesnumbered,ruled]{algorithm2e}
\usepackage{amssymb}
\usepackage{lineno}
\usepackage{float}
\usepackage{algorithmic}
\usepackage[tight,footnotesize]{subfigure}
\usepackage{booktabs}
\usepackage{caption}
\usepackage{makecell}
\usepackage{adjustbox}
\usepackage{wrapfig}
\usepackage{xcolor}
\usepackage[linesnumbered,ruled]{algorithm2e}
\usepackage{listings}
\usepackage{makecell}
\usepackage{pifont}
\usepackage{tikz}
\usetikzlibrary{arrows.meta, positioning, calc}

\newcommand{\cmark}{\ding{51}}%
\newcommand{\xmark}{\ding{55}}%
\newcommand{\pmark}{\ding{108}}       

\providecommand{\cmark}{\ding{51}}
\providecommand{\xmark}{\ding{55}}
\definecolor{vbslate}{RGB}{70,90,110}
\definecolor{vbamber}{RGB}{200,150,40}
\definecolor{vbteal}{RGB}{40,140,130}
\definecolor{vbbrick}{RGB}{180,70,60}

\usepackage{tikz}
\usetikzlibrary{arrows.meta,positioning,fit,backgrounds,calc,%
                shapes.geometric,decorations.pathreplacing}
 
\definecolor{mcNavy}{RGB}{29,53,87}      
\definecolor{mcBlue}{RGB}{69,123,157}    
\definecolor{mcLight}{RGB}{168,218,220}  
\definecolor{mcSand}{RGB}{241,235,222}   
\definecolor{mcRed}{RGB}{190,52,52}      
\definecolor{mcAmber}{RGB}{224,159,62}   
\definecolor{mcGreen}{RGB}{75,130,90}    
\definecolor{mcGrey}{RGB}{130,130,130}   

\definecolor{bcNavy}{RGB}{27,54,93}      
\definecolor{bcBlue}{RGB}{62,120,178}    
\definecolor{bcLight}{RGB}{173,216,222}  
\definecolor{bcSand}{RGB}{243,238,228}   
\definecolor{bcGreen}{RGB}{70,135,95}    
\definecolor{bcAmber}{RGB}{222,158,60}   
\definecolor{bcRed}{RGB}{188,60,55}      
\definecolor{bcGrey}{RGB}{130,130,130}   

\newcommand{\pv}[1]{\textcolor{mcRed}{#1}}   

\usepackage{pifont}
\providecommand{\cmark}{\ding{51}}
\providecommand{\xmark}{\ding{55}}
\providecommand{\pmark}{$\sim$}

\journal{Journal Name}

\begin{document}
\sloppy
\setlength{\parskip}{0pt}

\begin{frontmatter}

\title{BaseCamp --- An Agentic AI Framework for Automating DNA Sequencing Data Pipelines}


\author[label1]{Eranga Bandara}
\ead{cmedawer@odu.edu}

\author[label3]{Xueping Liang}
\ead{xuliang@fiu.edu}

\author[label10]{Asanga Gunaratna}
\ead{asanga.gunaratna@complianceoslab.app}

\author[label20]{Tharaka Hewa}
\ead{tharaka.hewa@oulu.fi}

\author[label3]{Abdul Rahman}
\ead{abdulrahman@deloitte.com}

\author[label1]{Peter Foytik}
\ead{pfoytik@odu.edu}

\author[label1]{Safdar H. Bouk}
\ead{sbouk@odu.edu}





\author[label7]{Sachini Rajapakse}
\ead{sachini.rajapakse@iciclelabs.ai}

\author[label7]{Isurunima Kularathna}
\ead{isurunima.kularathna@iciclelabs.ai}

\author[label7]{Pramoda Karunarathna}
\ead{pramoda.karunarathna@iciclelabs.ai}

\author[label7]{Chalani Rajapakse}
\ead{chalani.rajapakse@iciclelabs.ai}


\author[label5]{Ng Wee Keong}
\ead{awkng@ntu.edu.sg}

\author[label6]{Kasun De Zoysa}
\ead{kasun@ucsc.cmb.ac.lk}

\author[label8]{Amin Hass}
\ead{amin.hassanzadeh@accenture.com}


\author[label17]{Wathsala Herath}
\ead{wathsala.herath@agentsway.ai}


\author[label1]{Ross Gore}
\ead{rgore@odu.edu}

\author[label1]{Ravi Mukkamala}
\ead{rmukkama@odu.edu}



\author[label16]{Nihal Siriwardanagea}
\ead{nihal@gsi2.com}

\author[label18]{Gihan Siriwardanagea}
\ead{gihan.siriwardanagea@stud.lsmu.lt}


\author[label15]{Aruna Withanage}
\ead{aruna@effectz.ai}

\author[label15]{Nilaan Loganathan}
\ead{nilaan@effectz.ai}

\author[label1]{Sachin Shetty}
\ead{sshetty@odu.edu}

\address[label1]{Old Dominion University, Norfolk, VA, USA}
\address[label4]{Florida International University, USA}
\address[label10]{AI Motion Labs, Melbourne, Australia}
\address[label17]{Agentsway.AI}
\address[label3]{Deloitte \& Touche LLP, USA}
\address[label20]{Center for Wireless Communications, University of Oulu, Finland}
\address[label5]{Nanyang Technological University, Singapore}
\address[label6]{University of Colombo, Sri Lanka}
\address[label7]{IcicleLabs.AI}
\address[label8]{Accenture Technology Labs, Arlington, VA, USA}
\address[label16]{GSI Scandinavia AB}
\address[label18]{Lithuanian University of Health Sciences}
\address[label15]{Effectz.AI}

\begin{abstract}

DNA sequencing pipelines — spanning raw read quality control, reference alignment, variant calling, and annotation — are now reliably executed by workflow management systems that orchestrate established bioinformatics tools reproducibly at scale. What remains manual is the decision layer surrounding that execution: selecting quality thresholds appropriate to a given sample and platform, interpreting quality reports in context, adjudicating borderline variant calls, diagnosing anomalies, and determining which findings warrant expert review. These decisions are repetitive, judgment-intensive, inconsistently exercised across operators, and frequently undocumented. This paper introduces BaseCamp, a novel agentic AI framework for automating the decision layer of end-to-end DNA sequencing pipelines. The framework decomposes the pipeline into six specialized AI agents — covering sample intake and quality control, alignment, variant calling, annotation and interpretation, cross-stage monitoring, and reporting. Critically, BaseCamp agents do not perform sequence analysis: established, independently validated tools execute alignment, calling, and annotation, while the agents select among them, configure them, interpret their output, and decide what follows. This division confines language model reasoning to the judgment layer where it is reliable and preserves the reproducibility that existing tooling guarantees. To ensure contextual accuracy and adherence to responsible and explainable AI principles, the framework employs a consortium of fine-tuned, domain-specialized large language models coordinated by a central reasoning LLM, with all inference executing locally so that no sequencing data leaves the operating environment. A human-in-the-loop orchestration model allows bioinformaticians to supervise, validate, and intervene across pipeline stages via a Model Context Protocol (MCP)–enabled interface. Evaluation demonstrates that agent-generated configurations are concordant with expert practice, that an explicit filtering ledger renders inspectable what filtering otherwise removes without trace, and that cross-stage anomaly detection surfaces conditions — samples passing every individual stage check yet jointly implausible — that execution monitoring does not detect at all. Although grounded in a genomics operational environment, BaseCamp is domain-independent and offers a generalizable blueprint for agentic AI–driven automation of scientific data pipelines.

\end{abstract}

\begin{keyword}
Agentic AI \sep DNA Sequencing \sep Bioinformatics Pipeline Automation \sep
Variant Calling \sep Responsible AI \sep Explainable AI \sep LLM \sep Model Context Protocol
\end{keyword}

\end{frontmatter}

\section{Introduction}

DNA sequencing has become a foundational tool across genomics, agricultural
breeding, clinical diagnostics, and biomedical research, with sequencing
throughput increasing and per-genome cost falling steadily over the past
decade~\cite{wetterstrand2021dna, stephens2015big}. Converting raw sequencer
output into interpretable genetic information requires a multi-stage
computational pipeline spanning raw read quality control, adapter trimming,
reference alignment, duplicate marking, variant calling, filtering, and
functional annotation~\cite{gatk, deepvariant, bwa, samtools}. Mature, widely
adopted tools exist for each individual stage~\cite{fastqc, trimmomatic, vep,
gnomad}, and pipeline orchestration frameworks such as Nextflow and
Snakemake~\cite{nextflow, snakemake} enable these tools to be chained into
reproducible computational workflows. Despite this tooling maturity, the
decision-making layer surrounding pipeline execution --- selecting QC
thresholds appropriate to a given sample and sequencing platform, choosing
alignment and variant-calling parameters, interpreting borderline variant
calls, triaging pipeline failures, and prioritizing which findings warrant
expert review --- remains predominantly manual, dependent on bioinformatician
availability, and inconsistent across operators and sequencing runs.

The emergence of agentic AI systems represents a fundamental shift in how such
operational complexity can be addressed~\cite{sapkota2025, agent-survey}.
Unlike deterministic workflow managers that execute a fixed sequence of tool
invocations, agentic AI systems are capable of autonomous reasoning,
multi-step planning, and context-aware decision-making across interconnected
pipeline stages~\cite{agentic-ai-opptunities}. Networks of specialized AI
agents can continuously interpret quality control metrics, reason over
alignment statistics, evaluate variant call confidence in light of sequencing
depth and platform-specific error profiles, and flag anomalous runs for
review --- tasks that currently depend on the availability and judgment of an
experienced bioinformatician and are prone to inconsistency when performed
under time pressure across large sample batches~\cite{agentic-ai-rise}. By
delegating these reasoning-intensive responsibilities to autonomous agents
operating under human supervision, sequencing facilities and research groups
can achieve levels of throughput, consistency, and reproducibility that
manual oversight cannot sustain at scale.

Despite this potential, the application of agentic AI to DNA sequencing
pipelines remains largely unexplored in practice. Most existing applications
of large language models in genomics focus on isolated capabilities: genomic
language models that learn sequence-level representations directly from
DNA~\cite{dnabert, hyenadna}, or biomedical question-answering systems that
retrieve and summarize genomic literature~\cite{genegpt}, rather than
end-to-end reasoning over the operational pipeline that produces variant
calls in the first place. These point solutions improve specific sub-tasks
but do not address the coordination structure of the sequencing workflow
itself --- the sequence of decisions connecting raw reads to a reviewed,
reported variant set. The integration of multiple autonomous agents into a
cohesive, human-supervised pipeline capable of reasoning across the full
sequencing lifecycle, from raw read intake to annotated variant report,
represents an open challenge that existing agentic AI transition approaches
have not yet addressed in this domain~\cite{agentic-ai-transition-organization}.

A central barrier to this transition is the absence of a principled
framework for decomposing the sequencing pipeline into agent-executable
analytical roles, designing the coordination interfaces between agents and
existing bioinformatics tooling, and maintaining meaningful human oversight
at each decision point~\cite{agentic-ai-taxonomy-challenges,
agentic-workflow-practicle-guide}. Conventional pipeline automation is well
suited to deterministic, rule-bound execution, but is poorly suited to the
contextual judgment that pipeline operation frequently requires --- for
example, whether a borderline variant call warrants confidence given local
depth and mapping quality, or whether an unusual quality control profile
reflects a genuine sample issue rather than a benign platform artifact. At
the same time, sequencing pipelines routinely inform decisions with direct
downstream consequences --- clinical variant interpretation, breeding
selection, or diagnostic reporting --- where errors are costly and where
responsible AI design, explainability, and human-in-the-loop oversight are
essential requirements rather than optional enhancements~\cite{towards-rai-xai,
xai}.

In response to these challenges, this paper proposes BaseCamp, a novel
agentic AI framework for automating end-to-end DNA sequencing data pipelines.
BaseCamp introduces a multi-agent architecture in which the manual
decision-making layer of the sequencing pipeline is systematically decomposed
into specialized AI agents, each responsible for a clearly defined analytical
role~\cite{agentic-ai-workflow-patterns}. Agents are coordinated through a
human-in-the-loop orchestration model in which bioinformaticians supervise,
validate, and intervene at defined checkpoints via a Model Context Protocol
(MCP)--enabled interface~\cite{mcp1, mcp2, mcc}. To ensure contextual
accuracy and responsible, explainable decision-making, the framework employs
a consortium of fine-tuned, domain-specialized large language models
coordinated by a central reasoning LLM~\cite{reasoning-llms, gpt-oss},
enabling agents to reason jointly over pipeline logs, quality metrics,
alignment statistics, and variant call outputs. This architecture follows
the same consortium-plus-reasoning-LLM design applied in our prior work on
retail supply chain automation~\cite{flowr} and tourism SME workflow
automation~\cite{rovanima}, extending it to a domain in which the underlying
computation is dominated by established bioinformatics tools rather than by
natural-language business processes, and in which the value of agentic
reasoning lies in the judgment layer surrounding those tools rather than in
replacing them.

As a proof of concept, BaseCamp was implemented and evaluated using
real-world sequencing datasets. The sequencing operational environment
exhibits characteristics that are well suited to agentic AI automation,
including high sample throughput, well-defined but sample-dependent quality
thresholds, multiple sequential decision points at which contextual judgment
is required, and a strong requirement for reproducibility and audit trail ---
properties that render it a representative and rigorously challenging
validation environment for the proposed framework. The primary contributions
of this paper are fourfold:

\begin{enumerate}
    \item A novel agentic AI framework, BaseCamp, for end-to-end automation
    of DNA sequencing data pipelines, encompassing quality control,
    alignment, variant calling, annotation, and pipeline exception handling
    under a unified multi-agent architecture.

    \item A human-in-the-loop orchestration model in which bioinformaticians
    act as supervisors and decision-makers over autonomous agent workflows
    via MCP-enabled interfaces, ensuring accountability, transparency, and
    operational control throughout the sequencing lifecycle.

    \item A responsible and explainable AI design incorporating a consortium
    of fine-tuned, domain-specialized LLMs coordinated by a reasoning LLM,
    supporting explainable, context-aware, and reproducible decision-making
    across all agent interactions.

    \item A real-world proof of concept demonstrating the practical
    applicability of BaseCamp within a genomics sequencing operational
    environment, validating the framework's effectiveness in reducing manual
    pipeline configuration overhead and improving anomaly detection across
    sequencing runs.
\end{enumerate}

The remainder of this paper is organized as follows. Section 2 provides
background on key concepts underlying the BaseCamp framework, including
large language models, agentic AI systems, LLM consortiums, and
human-in-the-loop orchestration. Section 3 reviews related work in
bioinformatics pipeline automation, LLM-based agents for genomics, and
agentic AI workflow systems. Section 4 describes the manual sequencing
pipeline workflows that BaseCamp targets and establishes the operational
context. Section 5 presents the BaseCamp framework, detailing the
multi-agent architecture, agent roles, coordination model, and LLM
consortium design. Section 6 describes the proof-of-concept implementation
and evaluation of BaseCamp within a real-world sequencing operational
environment. Section 7 concludes the paper and outlines directions for
future research.

\section{Background}
\label{sec:background}

This section introduces the key concepts and technologies that underpin the
BaseCamp framework. Understanding these foundations is necessary before
examining how they are applied to automate DNA sequencing data pipelines in
subsequent sections.

\subsection{Large Language Models}
\label{subsec:llm}

Large Language Models (LLMs) are a class of artificial intelligence systems
trained on vast corpora of text data, enabling them to understand, generate,
and reason over natural language with a high degree of fluency and contextual
awareness~\cite{llm, gpt-llm}. Unlike earlier-generation models trained for
narrow, task-specific purposes, LLMs acquire broad general knowledge and
linguistic capability through exposure to diverse text sources, allowing them
to perform summarisation, question answering, structured data extraction, and
complex reasoning through natural language interaction
alone~\cite{llm-security, llm-agents}.

It is important to distinguish the role of LLMs in BaseCamp from that of
genomic language models. Models such as DNABERT and
HyenaDNA~\cite{dnabert, hyenadna} are trained directly on nucleotide sequences
and learn representations of the DNA sequence itself. BaseCamp does not
operate on nucleotide sequences. Its LLMs reason over the \textit{operational
artifacts} that a sequencing pipeline produces --- FastQC quality reports,
alignment summary statistics, duplication and coverage metrics, variant call
format (VCF) records, annotation tables, and tool logs~\cite{fastqc, samtools,
vcfspec}. These artifacts are predominantly semi-structured text and tabular
summaries accompanied by tool-generated narrative output, and the capacity to
interpret such heterogeneous input and emit structured, actionable decisions
is precisely what makes agentic reasoning tractable in this domain. The
computation itself remains the responsibility of established bioinformatics
tools; the LLM supplies the judgment layer around them.

\subsection{Reasoning LLMs}
\label{subsec:reasoningllm}

While standard LLMs excel at language understanding and generation, reasoning
LLMs extend these capabilities by incorporating explicit chain-of-thought
processes, multi-step planning, and self-verification mechanisms that enable
more reliable and structured decision-making~\cite{reasoning-llms,
llm-reasoning}. Reasoning LLMs are designed to evaluate intermediate
conclusions, identify inconsistencies across multiple sources of evidence, and
synthesise coherent final outputs from complex, multi-perspective inputs.

In the BaseCamp framework, a central reasoning LLM --- specifically OpenAI
GPT-OSS~\cite{gpt-oss} --- serves as the supervisory model within the LLM
consortium architecture. When multiple fine-tuned domain-specialised LLMs
produce independent outputs in response to an agent's prompt, the reasoning
LLM evaluates, compares, and synthesises these responses to generate a final
decision or recommendation. This ensemble-based reasoning approach reduces the
risk of model-specific bias or error propagation, and ensures that automated
pipeline decisions are validated across multiple reasoning perspectives before
execution. This requirement is acute in a sequencing context: a decision to
apply a hard quality filter, to accept a borderline variant call, or to pass a
marginal sample through to downstream analysis propagates silently into every
result derived from it, and is frequently not revisited once made.

\subsection{LLM Fine-Tuning}
\label{subsec:finetuning}

While pre-trained LLMs possess broad general knowledge, their performance on
domain-specific tasks --- particularly those requiring precise operational
vocabulary, structured output formats, and context-aware reasoning within a
specific technical domain --- can be significantly improved through
fine-tuning~\cite{mistral-fine-tune, llm-finetune}. Fine-tuning adapts a
general-purpose pre-trained base model on a curated, domain-specific dataset
through supervised training, enabling the model to internalise domain
conventions, terminologies, and reasoning patterns that are not well
represented in general pre-training corpora. Genomics is a strong candidate
for such adaptation: quality thresholds, filter expressions, VCF field
semantics, and platform-specific error profiles constitute a dense technical
vocabulary that a general model handles inconsistently.

\begin{figure}[H]
\centering
\resizebox{\textwidth}{!}{%
\begin{tikzpicture}[
  font=\sffamily\scriptsize,
  src/.style={draw=bcGrey!55,fill=white,rounded corners=3pt,text width=34mm,
              align=center,minimum height=10mm,inner sep=3pt,font=\sffamily\scriptsize},
  proc/.style={draw=bcBlue,fill=bcSand,rounded corners=3pt,text width=36mm,
               align=center,minimum height=12mm,inner sep=3pt,font=\sffamily\scriptsize},
  basebox/.style={draw=bcNavy,fill=bcNavy,text=white,rounded corners=3pt,text width=30mm,
               align=center,minimum height=12mm,inner sep=3pt,font=\sffamily\scriptsize\bfseries},
  outbox/.style={draw=bcRed!75,fill=bcRed!7,rounded corners=3pt,text width=32mm,
              align=center,minimum height=12mm,inner sep=3pt,font=\sffamily\scriptsize},
  band/.style={rounded corners=4pt,draw=bcGrey!30,fill=bcGrey!4,line width=0.4pt},
  hdr/.style={font=\sffamily\scriptsize\bfseries,text=bcNavy,anchor=west},
  note/.style={font=\sffamily\tiny\itshape,text=bcGrey},
  fl/.style={draw=bcBlue,-{Latex[length=1.6mm]},line width=0.5pt},
  flg/.style={draw=bcGrey!55,-{Latex[length=1.5mm]},line width=0.45pt}
]

\draw[band] (-8.5,2.6) rectangle (8.5,0.55);    
\draw[band] (-8.5,0.15) rectangle (8.5,-2.35);  
\draw[band] (-8.5,-2.75) rectangle (8.5,-5.0);  

\node[hdr] at (-8.3,2.3)  {SEQUENCING OPERATIONS CORPUS};
\node[hdr] at (-8.3,-0.15) {SUPERVISED FINE-TUNING};
\node[hdr] at (-8.3,-3.05) {LOCAL DEPLOYMENT};

\node[src] (d1) at (-6.0,1.45) {QC reports\\[1pt]{\tiny\itshape\color{bcGrey} + dispositions}};
\node[src] (d2) at (-2.0,1.45) {pipeline logs\\[1pt]{\tiny\itshape\color{bcGrey} + parameter settings}};
\node[src] (d3) at ( 2.0,1.45) {filtering decisions\\[1pt]{\tiny\itshape\color{bcGrey} + recorded rationale}};
\node[src] (d4) at ( 6.0,1.45) {review notes\\[1pt]{\tiny\itshape\color{bcGrey} bioinformatician}};

\node[basebox] (bm)  at (-6.0,-1.3) {Base LLM\\[1pt]{\tiny\itshape\mdseries Llama-3 $\cdot$ Mistral $\cdot$ Qwen\,2.5}};
\node[proc]    (cur) at (-1.7,-1.3) {curation \& split\\[1pt]{\tiny\itshape\color{bcGrey} 80 / 10 / 10}};
\node[proc]    (ft)  at ( 2.6,-1.3) {fine-tuning\\[1pt]{\tiny\itshape\color{bcGrey} Unsloth $\cdot$ LoRA + QLoRA}};
\node[outbox]  (fm)  at ( 6.6,-1.3) {domain adapter\\[1pt]{\tiny\itshape\color{bcGrey} one per function}};

\foreach \d/\x in {d1/-6.0, d2/-2.0, d3/2.0, d4/6.0}{%
  \draw[flg] (\d.south) -- (\x,0.35);}
\draw[draw=bcGrey!55,line width=0.45pt] (-6.0,0.35) -- (6.0,0.35);
\draw[flg] (-1.7,0.35) -- (cur.north);

\draw[fl] (bm.east)  -- (cur.west);
\draw[fl] (cur.east) -- (ft.west);
\draw[fl] (ft.east)  -- (fm.west);

\node[proc,draw=bcGreen!80,fill=bcGreen!10] (ol) at (-4.4,-4.0)
  {Ollama local serving\\[1pt]{\tiny\itshape\color{bcGrey} hot-swap adapter per agent}};
\node[src] (inf) at (0.4,-4.0)
  {low-latency inference\\[1pt]{\tiny\itshape\color{bcGrey} no external API dependency}};
\node[src,draw=bcGreen!80] (gov) at (5.2,-4.0)
  {data stays on-site\\[1pt]{\tiny\itshape\color{bcGrey} no sequencing data egress}};

\draw[fl] (fm.south) -- (6.6,-2.55) -- (-4.4,-2.55) -- (ol.north);
\draw[fl] (ol.east)  -- (inf.west);
\draw[fl] (inf.east) -- (gov.west);

\node[anchor=west,font=\sffamily\tiny\itshape,text=bcGrey,text width=160mm] at (-8.3,-4.8)
  {Fine-tuning is performed offline; deployed model behaviour is fixed during
   pipeline execution, so two runs of the same sample are reconcilable.};
\end{tikzpicture}}
\caption{Supervised fine-tuning pipeline used to adapt a base LLM to
sequencing pipeline operations. A curated corpus of QC reports, pipeline logs
and parameter settings, variant filtering decisions, and bioinformatician
review notes is used to fine-tune a base model, with Low-Rank Adapter (LoRA)
modules applied for parameter-efficient training under 4-bit
quantisation~\cite{lora, qlora}. The resulting adapter retains the general
linguistic capability of the base model while internalising sequencing-domain
vocabulary and reasoning patterns, and is served locally via
Ollama~\cite{ollama} so that no sequencing data leaves the operating
environment.}
\label{fig:finetune}
\end{figure}
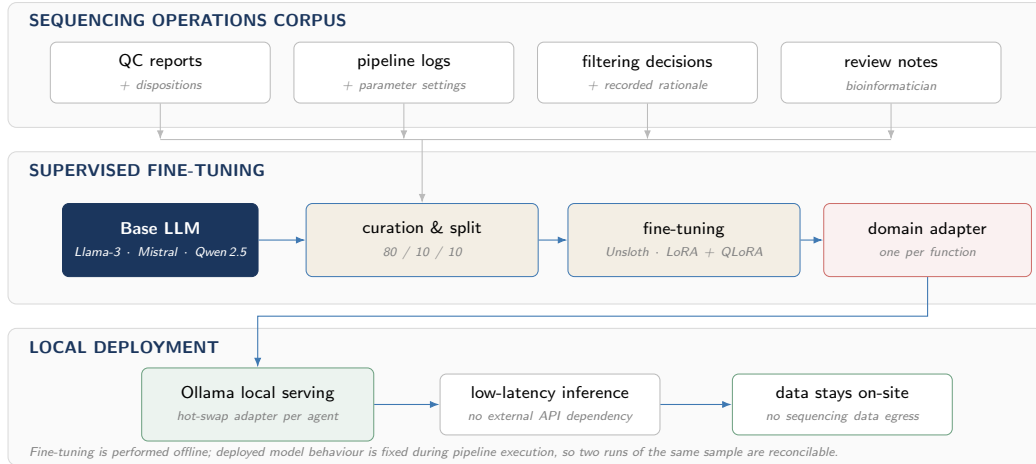

As illustrated in Figure~\ref{fig:finetune}, the fine-tuning pipeline employed
in BaseCamp begins with a base LLM --- such as Llama-3, Mistral, or
Qwen~\cite{llama-3, mistral-llm, qwen2} --- which is adapted using a curated
sequencing operations dataset comprising historical QC reports, pipeline
execution logs, parameter configurations paired with the outcomes they
produced, variant filtering decisions, and bioinformatician review notes. To
enable parameter-efficient training within resource-constrained environments,
Low-Rank Adapters (LoRA)~\cite{lora} are applied during fine-tuning, and
4-bit quantisation (QLoRA)~\cite{qlora} further reduces memory requirements.
Fine-tuned models are deployed locally using Ollama~\cite{ollama}, enabling
low-latency agent inference without dependence on external API services --- a
property that also matters where sequencing data is subject to institutional
or regulatory restrictions on external transmission.

\subsection{AI Agents and Agentic AI}
\label{subsec:agenticai}

Traditional LLM interactions follow a simple request--response pattern in
which a human constructs a prompt, submits it to a language model, and
manually interprets the response to determine subsequent actions. In this
paradigm the human remains responsible for orchestration, decision-making, and
follow-up execution, with the LLM functioning as a passive assistant rather
than an active participant in workflow execution~\cite{agentic-ai,
agentic-workflow-practicle-guide}.

\begin{figure}[H]
\centering
\resizebox{\textwidth}{!}{%
\begin{tikzpicture}[
  font=\sffamily\scriptsize,
  hum/.style={draw=bcGreen!80,fill=bcGreen!12,rounded corners=3pt,text width=28mm,
              align=center,minimum height=12mm,inner sep=3pt,font=\sffamily\scriptsize},
  llmb/.style={draw=bcNavy,fill=bcNavy,text=white,rounded corners=3pt,text width=24mm,
               align=center,minimum height=11mm,inner sep=3pt,font=\sffamily\scriptsize\bfseries},
  stepbox/.style={draw=bcGrey!55,fill=white,rounded corners=3pt,text width=32mm,
               align=center,minimum height=12mm,inner sep=3pt,font=\sffamily\scriptsize},
  ag/.style={draw=bcBlue,fill=bcSand,rounded corners=3pt,text width=34mm,
             align=center,minimum height=26mm,inner sep=3.5pt,font=\sffamily\scriptsize},
  tool/.style={draw=bcGrey!55,fill=white,rounded corners=3pt,text width=32mm,
               align=center,minimum height=8mm,inner sep=2.5pt,font=\sffamily\scriptsize},
  pnl/.style={rounded corners=5pt,draw=bcGrey!35,fill=bcGrey!4,line width=0.5pt},
  hdr/.style={font=\sffamily\scriptsize\bfseries,text=bcNavy,anchor=west},
  note/.style={font=\sffamily\tiny\itshape,text=bcGrey},
  fl/.style={draw=bcBlue,-{Latex[length=1.7mm]},line width=0.55pt},
  flt/.style={draw=bcGrey!60,{Latex[length=1.5mm]}-{Latex[length=1.5mm]},line width=0.45pt},
  flg/.style={draw=bcGreen!80,-{Latex[length=1.7mm]},line width=0.55pt}
]

\draw[pnl] (-8.6,1.9) rectangle (8.6,-2.6);
\node[hdr] at (-8.35,1.55) {(a) \ DIRECT HUMAN--LLM INTERACTION};

\node[hum]     (h1) at (-6.2,0.15) {\textbf{Bioinformatician}};
\node[stepbox] (m1) at (-2.0,0.15) {construct prompt\\[1.5pt]{\tiny\itshape\color{bcGrey} paste QC report / VCF excerpt}};
\node[llmb]    (l1) at ( 1.9,0.15) {LLM};
\node[stepbox] (r1) at ( 6.0,0.15) {interpret response\\[1.5pt]{\tiny\itshape\color{bcGrey} and act manually}};

\draw[fl] (h1.east) -- (m1.west);
\draw[fl] (m1.east) -- (l1.west);
\draw[fl] (l1.east) -- (r1.west);

\draw[flg,dashed] (r1.south) -- (6.0,-1.55) -- (-6.2,-1.55) -- (h1.south);
\node[note,text=bcGreen!80,anchor=north] at (-0.1,-1.72)
  {the human performs every orchestration step};

\draw[pnl] (-8.6,-3.3) rectangle (8.6,-11.3);
\node[hdr] at (-8.35,-3.7) {(b) \ AGENTIC AI--LLM INTERACTION};

\node[llmb] (l2) at (-1.9,-4.85) {LLM};

\node[hum] (h2) at (-6.2,-7.3) {\textbf{Bioinformatician}\\[1.5pt]{\tiny\itshape\color{bcGrey} supervisor}};

\node[ag] (a2) at (-1.9,-7.3)
  {\textbf{AI Agent}\\[3pt]
   {\tiny construct prompt}\\[2pt]
   {\tiny invoke model}\\[2pt]
   {\tiny interpret response}\\[2pt]
   {\tiny invoke tool}\\[2pt]
   {\tiny decide next action}};

\node[tool] (t1) at (6.0,-5.5)  {FastQC $\cdot$ MultiQC};
\node[tool] (t2) at (6.0,-6.7)  {BWA $\cdot$ SAMtools};
\node[tool] (t3) at (6.0,-7.9)  {GATK $\cdot$ DeepVariant};
\node[tool] (t4) at (6.0,-9.1)  {VEP $\cdot$ annotation DBs};

\draw[flg] (h2.east) -- (a2.west);
\node[note,text=bcGreen!80,anchor=south] at (-4.05,-7.2) {goal};

\draw[flg,dashed] (a2.south) -- (-1.9,-10.1) -- (-6.2,-10.1) -- (h2.south);
\node[note,text=bcGreen!80,anchor=north] at (-4.05,-10.27) {approve / override};

\draw[fl,{Latex[length=1.6mm]}-{Latex[length=1.6mm]}] (a2.north) -- (l2.south);
\node[note,text=bcNavy,anchor=west] at (-1.72,-5.75) {reason};

\draw[flt] (a2.east) -- (2.4,-7.3);
\draw[draw=bcGrey!60,line width=0.45pt] (2.4,-5.5) -- (2.4,-9.1);
\foreach \t/\y in {t1/-5.5, t2/-6.7, t3/-7.9, t4/-9.1}{%
  \draw[flt] (2.4,\y) -- (\t.west);}
\node[note,anchor=south] at (3.2,-7.2) {invoke \ / \ return};
\node[note,anchor=north] at (6.0,-9.65) {established tools compute};

\node[anchor=west,font=\sffamily\tiny\itshape,text=bcGrey,text width=150mm] at (-8.35,-11.0)
  {The agent decides; it does not compute. Sequence analysis remains the
   responsibility of established, independently validated tools.};
\end{tikzpicture}}
\caption{Comparison between direct human--LLM interaction and agentic AI--LLM
interaction in a sequencing context. In panel (a) the bioinformatician
constructs each prompt, submits it to the model, and manually interprets and
acts on the response, remaining responsible for every orchestration step. In
panel (b) an AI agent autonomously manages the interaction loop --- prompt
construction, model invocation, response interpretation, external tool
invocation, and determination of subsequent actions --- while the
bioinformatician supplies the goal and retains approval authority. Crucially,
the agent does not itself perform sequence analysis: it invokes established
bioinformatics tools~\cite{fastqc, bwa, samtools, gatk, deepvariant, vep} and
reasons over their output, preserving the reproducibility those tools already
guarantee.}
\label{fig:aiagent}
\end{figure}
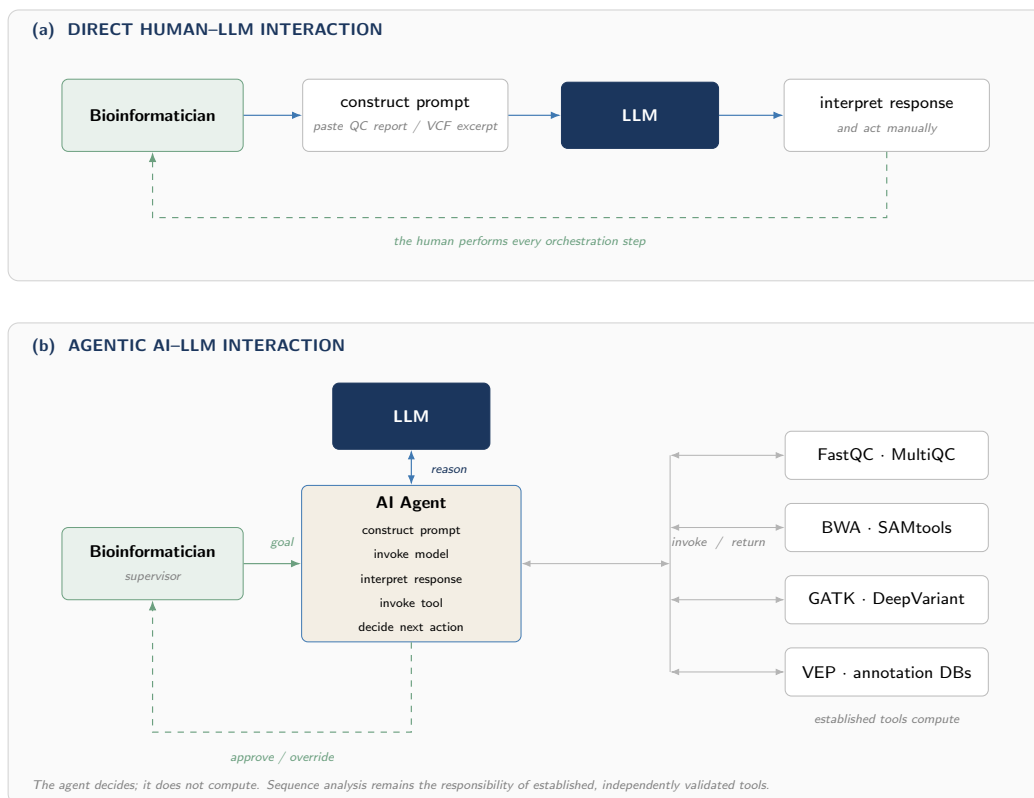

An AI agent fundamentally changes this interaction model. As illustrated in
Figure~\ref{fig:aiagent}, an AI agent autonomously manages the full
interaction loop --- constructing prompts, invoking language models,
interpreting responses, invoking external tools, and determining subsequent
actions --- without direct human intervention at each step. AI agents are
software systems that leverage LLMs in combination with tools, APIs, memory,
and external context to execute tasks iteratively and
autonomously~\cite{agentsway}. When multiple specialised agents collaborate,
each assigned distinct responsibilities, they form \textit{agentic AI
workflows}~\cite{sapkota2025, agent-survey,
agentic-ai-transition-organization}.

The distinction matters for sequencing because the tool-invocation capability
is not incidental --- it is the mechanism by which agentic reasoning is
grounded in real computation. A BaseCamp agent does not estimate alignment
quality; it invokes SAMtools or a comparable tool~\cite{samtools, bwa},
reasons over the returned statistics, and decides what follows. This
separation --- established tools compute, agents decide --- distinguishes the
framework from approaches that ask a language model to perform analysis it is
not suited to, and it preserves the reproducibility guarantees that existing
bioinformatics tooling already provides.

\subsection{Model Context Protocol}
\label{subsec:mcp}

The Model Context Protocol (MCP) is a standardised, open protocol defining how
AI agents connect to and interact with external systems, tools, databases, and
APIs through secure, structured interfaces~\cite{mcp1, mcp2, mcc}. MCP enables
agents to access real-time data and invoke external capabilities --- such as
launching an alignment job, querying a variant annotation database, or
retrieving run metadata from a laboratory information management system
(LIMS) --- in a modular and interoperable manner, without requiring custom
integration code for each system connection.

In BaseCamp, each agentic workflow is exposed as an independent MCP server,
enabling a single bioinformatician to connect to and orchestrate multiple
agent workflows simultaneously through a unified natural language interface
via LM Studio~\cite{lm-studio}. This architecture decouples agent logic from
the underlying systems it drives: the QC Agent reaches FastQC and MultiQC
through its own MCP server, the Alignment Agent reaches BWA and SAMtools
through another, the Variant Calling Agent the calling and filtering stack
through a third, and so on~\cite{fastqc, multiqc, bwa, samtools, gatk}.
Individual workflows can therefore evolve independently --- a variant caller
can be upgraded, or a filtering policy revised, without disrupting the broader
system. Because servers are individually addressable, a bioinformatician who
disputes a single stage can re-run that stage alone rather than recomputing
the entire pipeline, which is the common case when a run is questioned in part
rather than in whole, and a material advantage where full re-execution is
measured in hours. The resulting deployment topology is described in
Section~\ref{subsec:impl_overview}.

\subsection{Responsible and Explainable AI}
\label{subsec:responsibleai}

As agentic AI systems take on increasingly consequential roles in scientific
data pipelines --- setting quality thresholds, filtering variant calls, and
determining which findings reach human review --- ensuring that their
decisions are transparent, accountable, and reproducible becomes a
foundational requirement rather than an optional
enhancement~\cite{responsible-ai, xai}. Responsible AI encompasses fairness,
transparency, accountability, privacy, and human oversight in the design and
deployment of AI systems. Explainable AI (XAI) refers specifically to the
capacity of a system to provide human-interpretable justification for its
outputs --- enabling operators to understand not merely what was decided, but
why~\cite{towards-rai-xai, llm-explainability}.

The stakes in a sequencing context are distinctive in two respects. First,
pipeline decisions are \textit{silent and cumulative}: a filter applied at the
variant calling stage removes calls that no downstream analysis will ever see,
and the absence leaves no trace unless it is deliberately recorded. Second,
sequencing results routinely inform consequential downstream decisions ---
clinical variant interpretation, breeding selection, or diagnostic
reporting --- where an error propagates well beyond the pipeline that produced
it. Reproducibility is therefore not merely good practice but a precondition
for the scientific validity of anything derived from the
output~\cite{nextflow, snakemake}.

BaseCamp operationalises these principles through four complementary
mechanisms. First, the LLM consortium architecture ensures that consequential
decisions are validated across multiple independent reasoning perspectives
before execution, reducing the risk of model-specific bias or hallucination.
Second, every agent output is structured to include an explicit reasoning
trace: each QC pass or fail cites the specific metric and threshold that
determined it, each parameter selection records its justification, and each
variant filtering decision records the criteria applied and the number of
calls affected. Third, human approval gates are embedded at defined decision
points, so that no consequential action --- such as committing a filtered
variant set to downstream analysis --- proceeds without explicit validation.
Fourth, the full configuration under which a run was executed --- tool
versions, parameters, model versions, and adapter revisions --- is recorded
with the output, so that any result can be reproduced or recomputed under
different settings. Together these mechanisms ensure that BaseCamp delivers
pipeline automation while preserving the auditability and reproducibility that
scientific use demands.

\section{Related Works}
\label{sec:related}

Research relevant to the BaseCamp framework spans four interconnected areas:
bioinformatics workflow management and pipeline automation, LLM-based agents
for genomics and biomedical analysis, agentic AI for scientific discovery and
autonomous experimentation, and agentic AI and multi-agent workflow systems
more generally. The following subsections review representative works in each
area and identify the gaps that BaseCamp addresses.

\subsection{Bioinformatics Workflow Management and Pipeline Automation}
\label{subsec:rw_workflow}

The dominant approach to sequencing pipeline automation is the workflow
management system. Nextflow~\cite{nextflow} and Snakemake~\cite{snakemake}
allow bioinformaticians to express a pipeline as a directed graph of tool
invocations with declared inputs, outputs, and dependencies, and to execute
that graph reproducibly across heterogeneous compute environments through
containerisation and portable execution backends. Galaxy~\cite{galaxy}
provides equivalent capability through a web-based graphical interface aimed
at users without command-line expertise, and the Common Workflow
Language~\cite{cwl} supplies a vendor-neutral specification for describing
such workflows portably. Building on these engines,
nf-core~\cite{nfcore} curates a community-maintained collection of
peer-reviewed, standardised pipelines --- including germline and somatic
variant calling workflows --- that encode established best practice and
substantially reduce the effort required to deploy a correct pipeline.
Reporting tools such as MultiQC~\cite{multiqc} aggregate quality metrics
across samples and pipeline stages into consolidated reports intended for
human inspection.

Collectively these systems solve the \textit{execution} problem, and solve it
well: given a specified pipeline and a specified parameter set, they run it
reproducibly, at scale, with full provenance. What they do not address is the
\textit{decision} problem that surrounds execution. A workflow manager will
faithfully execute whatever quality threshold it is given, but it does not
reason about whether that threshold is appropriate for a particular sample,
library preparation, or sequencing platform. It will emit a MultiQC report,
but it does not interpret one. It will fail a job, but it does not diagnose
why, nor decide whether the correct response is re-running with adjusted
parameters, flagging the sample for exclusion, or escalating to a human.
These judgments remain the responsibility of the bioinformatician, are
exercised inconsistently across operators, and constitute precisely the layer
BaseCamp targets. BaseCamp is therefore complementary to rather than
competitive with these systems: agents invoke established tooling and, where
appropriate, existing workflow engines, and supply the reasoning layer above
them.

\subsection{LLM-Based Agents for Genomics and Biomedical Analysis}
\label{subsec:rw_llmgenomics}

A rapidly growing body of work applies large language models to genomic and
biomedical tasks. One strand trains language models directly on nucleotide
sequences: DNABERT~\cite{dnabert} and HyenaDNA~\cite{hyenadna} learn
representations of DNA as a language, supporting downstream tasks such as
regulatory element prediction and variant effect estimation. This strand is
orthogonal to the present work --- these models operate on the sequence
itself, whereas BaseCamp operates on the operational artifacts a pipeline
produces, and the two could in principle be composed, with a genomic language
model invoked as a tool by an interpretation agent.

A second strand equips general-purpose LLMs with biomedical tools and data
access. GeneGPT~\cite{genegpt} augments an LLM with NCBI Web APIs to answer
genomics questions, demonstrating that tool augmentation substantially
outperforms parametric recall for factual genomic queries.
BioMANIA~\cite{biomania} translates natural language instructions into
executable calls against bioinformatics Python libraries, lowering the barrier
to tool use for non-programmers. AutoBA~\cite{autoba} advances further toward
autonomy, proposing an LLM-based agent that plans and executes multi-omic
analyses end-to-end from a description of the input data and analytical goal,
and CellAgent~\cite{cellagent} applies a multi-agent decomposition
--- planner, executor, evaluator --- to single-cell RNA sequencing analysis
tasks.

This strand is the closest prior art, and AutoBA and CellAgent in particular
establish the feasibility on which BaseCamp builds: LLM agents can plan and
execute real bioinformatics analyses at usable quality. Three differences
define the gap. First, these systems target \textit{downstream analysis}
--- differential expression, cell type annotation, multi-omic integration ---
taking a processed data matrix as their starting point, whereas BaseCamp
targets the \textit{upstream production pipeline} that generates such data
from raw reads, where the decisions concern quality thresholds, alignment
parameters, and variant filtering. Second, they rely predominantly on
general-purpose frontier models accessed via external APIs, with no
domain-specific fine-tuning and no provision for local execution --- a
constraint that is disqualifying where sequencing data is subject to
institutional or regulatory restriction on external transmission. Third, they
emphasise autonomy and task completion, with limited attention to structured
human approval gates, per-decision reasoning traces, or the configuration
recording that reproducible scientific use requires.

\subsection{Agentic AI for Scientific Discovery and Autonomous Experimentation}
\label{subsec:rw_science}

Beyond genomics, agentic AI has been applied to scientific work more broadly.
Coscientist~\cite{coscientist} demonstrates an LLM-driven system that
autonomously plans, designs, and executes chemical experiments using
laboratory robotics, and ChemCrow~\cite{chemcrow} augments an LLM with
expert-designed chemistry tools to perform synthesis planning and reaction
prediction, in both cases showing that tool-augmented agents substantially
outperform the underlying model operating alone. The AI Scientist~\cite{aiscientist}
extends the paradigm to the research process itself, generating hypotheses,
running experiments, and drafting manuscripts autonomously.

These works establish the broader premise --- that agentic decomposition with
tool augmentation is effective for technical scientific work --- and they also
surface the concerns that motivate BaseCamp's design constraints. Each
emphasises autonomous capability, and the accompanying discussions
consistently identify oversight, verifiability, and misuse as the open
problems rather than raw capability. BaseCamp occupies a deliberately
narrower operating envelope: it does not generate hypotheses, design
experiments, or produce scientific conclusions. It automates the decision
layer of an established, well-characterised production pipeline whose correct
operation is already defined by community best practice~\cite{nfcore, gatk},
and it does so under explicit human approval gates. The value claimed is
consistency and throughput in a known process, not autonomous discovery.

\subsection{Agentic AI and Multi-Agent Workflow Systems}
\label{subsec:rw_agentic}

Sapkota et al.~\cite{sapkota2025} provide a conceptual taxonomy
distinguishing AI agents from agentic AI systems, clarifying the architectural
and behavioural differences between single-agent tools and multi-agent
systems capable of sustained reasoning and coordinated action, and
identifying orchestration, role decomposition, and inter-agent communication
as the principal design challenges. Their contribution is conceptual and does
not address application to a specific operational domain. Broader surveys of
LLM-based autonomous agents~\cite{agent-survey, llm-agents} similarly
characterise the design space --- planning, memory, tool use, multi-agent
coordination --- without prescribing a transition path from an existing manual
process to an agentic one.

Bandara et al.~\cite{agentsway} introduce AgentsWay, a development
methodology for teams building agentic AI systems, emphasising AI-assisted
development, rapid iteration, and domain-driven workflow design. BaseCamp
follows this methodology directly. Prior applications of the same
consortium-plus-reasoning-LLM architecture have targeted retail supply chain
automation~\cite{flowr}, tourism SME operations~\cite{rovanima}, peer-based
mental health support in resource-constrained
environments~\cite{yarlagadda2026trainers}, and false memory risk assessment
in investigative and legal contexts~\cite{memcached}, in each case
demonstrating that decomposition into specialised agents under human-in-the-loop
orchestration is effective where a manual process is coordination-heavy and
judgement-intensive~\cite{agentic-workflow-practicle-guide,
agentic-ai-transition-organization}.

BaseCamp applies this architecture to a domain with a distinguishing property.
In the operational applications cited, agents both reason and act: the
consortium synthesises toward an action --- a purchase order, an itinerary,
a replenishment plan --- and the agent's output largely constitutes the work
product. In a sequencing pipeline the computation is performed by established,
independently validated tools~\cite{bwa, samtools, gatk, deepvariant, vep},
and the agent's contribution is confined to the decisions surrounding
them --- which tool, which parameters, which threshold, what to do when a stage
behaves unexpectedly, and what warrants human attention. This division is
deliberate and load-bearing: it preserves the reproducibility guarantees that
the existing tooling already provides, and it keeps the agent's role within
the bounds where LLM reasoning is reliable.

\subsection{Positioning}
\label{subsec:rw_positioning}

The pattern across the reviewed literature is consistent. Workflow management
systems solve reproducible execution but not the decisions that configure it.
Genomic language models operate on sequence rather than on pipeline
operations. LLM agents for bioinformatics target downstream analysis rather
than the upstream production pipeline, and generally depend on external
frontier-model APIs without domain fine-tuning or local execution. Agentic
systems for scientific discovery demonstrate the paradigm but prioritise
autonomy over the oversight and reproducibility that production pipelines
require. And general agentic AI frameworks supply the architectural pattern
without application to sequencing.

BaseCamp is distinguished not by any single column but by their conjunction:
end-to-end coverage of the sequencing production pipeline from raw read intake
to annotated variant report, decomposed across specialised agents that invoke
established bioinformatics tooling rather than replacing it, powered by a
consortium of locally served fine-tuned models with a reasoning layer, under
human approval gates at every consequential decision point, with full
configuration recording so that any run is reproducible. Table~\ref{tab:relatedwork}
presents a comparative analysis of the reviewed works in relation to the
proposed framework.

\begin{table}[H]
\centering
\caption{Comparison of related works and the BaseCamp framework.
\cmark~=~supported; \xmark~=~not supported; $\sim$~=~partially supported.}
\label{tab:relatedwork}
\begin{adjustbox}{width=\textwidth}
\begin{tabular}{lcccccccc}
\toprule
\thead{Work} &
\thead{Domain} &
\thead{Sequencing\\pipeline\\focus} &
\thead{End-to-end\\workflow} &
\thead{Agentic\\multi-agent} &
\thead{LLM\\based} &
\thead{LLM\\fine-tuning} &
\thead{Human-in-\\the-loop} &
\thead{Local /\\air-gapped\\execution} \\
\midrule
\textbf{BaseCamp (ours)} & \makecell{DNA sequencing\\pipeline automation}
& \cmark & \cmark & \cmark & \cmark & \cmark & \cmark & \cmark \\
\midrule
Nextflow~\cite{nextflow}        & \makecell{Workflow\\management}        & \cmark & \cmark & \xmark & \xmark & \xmark & \pmark & \cmark \\
Snakemake~\cite{snakemake}      & \makecell{Workflow\\management}        & \cmark & \cmark & \xmark & \xmark & \xmark & \pmark & \cmark \\
Galaxy~\cite{galaxy}            & \makecell{Workflow\\platform}          & \cmark & \cmark & \xmark & \xmark & \xmark & \cmark & \pmark \\
nf-core~\cite{nfcore}           & \makecell{Curated genomics\\pipelines} & \cmark & \cmark & \xmark & \xmark & \xmark & \pmark & \cmark \\
DNABERT~\cite{dnabert}          & \makecell{Genomic language\\model}     & \xmark & \xmark & \xmark & \cmark & \cmark & \xmark & \cmark \\
HyenaDNA~\cite{hyenadna}        & \makecell{Genomic sequence\\modelling} & \xmark & \xmark & \xmark & \cmark & \cmark & \xmark & \cmark \\
GeneGPT~\cite{genegpt}          & \makecell{Genomics question\\answering}& \xmark & \xmark & \xmark & \cmark & \xmark & \xmark & \xmark \\
BioMANIA~\cite{biomania}        & \makecell{NL-driven\\tool invocation}  & \pmark & \xmark & \xmark & \cmark & \xmark & \pmark & \xmark \\
AutoBA~\cite{autoba}            & \makecell{Automated multi-omic\\analysis} & \pmark & \pmark & \pmark & \cmark & \xmark & \pmark & \xmark \\
CellAgent~\cite{cellagent}      & \makecell{Single-cell RNA-seq\\analysis} & \xmark & \pmark & \cmark & \cmark & \xmark & \xmark & \xmark \\
Coscientist~\cite{coscientist}  & \makecell{Autonomous chemical\\experimentation} & \xmark & \cmark & \pmark & \cmark & \xmark & \pmark & \xmark \\
ChemCrow~\cite{chemcrow}        & \makecell{LLM chemistry\\tool agent}   & \xmark & \pmark & \xmark & \cmark & \xmark & \pmark & \xmark \\
Sapkota et al.~\cite{sapkota2025} & \makecell{Agentic AI\\taxonomy}      & \xmark & \xmark & \cmark & \cmark & \xmark & \xmark & \xmark \\
Bandara et al.~\cite{agentsway} & \makecell{Agentic AI\\methodology}     & \xmark & \xmark & \cmark & \cmark & \cmark & \cmark & \cmark \\
Flowr~\cite{flowr}              & \makecell{Retail supply chain\\automation} & \xmark & \cmark & \cmark & \cmark & \cmark & \cmark & \cmark \\
Rovanima~\cite{rovanima}        & \makecell{Tourism SME\\automation}     & \xmark & \cmark & \cmark & \cmark & \cmark & \cmark & \cmark \\
\bottomrule
\end{tabular}
\end{adjustbox}
\end{table}

\section{Manual DNA Sequencing Pipeline Workflows}
\label{sec:manual}

A successful agentic AI transition begins with a thorough understanding of the
manual processes that automation intends to replace. In the context of DNA
sequencing operations, the pipeline function encompasses a set of recurring,
decision-intensive stages spanning raw read quality assessment, reference
alignment, variant calling, functional annotation, and run-level monitoring
across sample batches~\cite{gatk, nfcore}. These stages are deeply
interdependent --- a permissive trimming decision propagates into alignment
statistics, which in turn condition variant call confidence --- yet in
practice each is configured and reviewed as a discrete step, frequently by
different personnel, with the connecting judgments recorded informally if at
all.

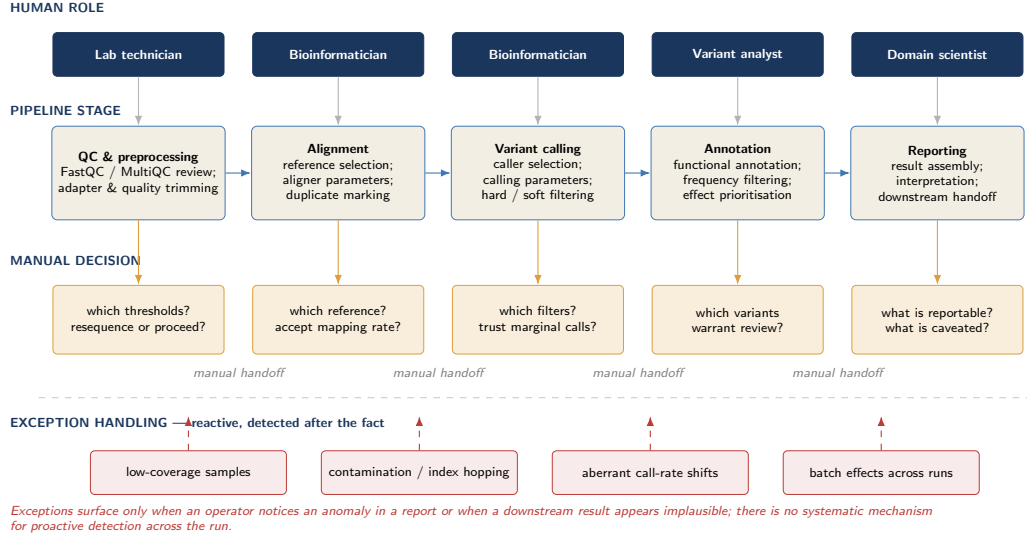
\begin{figure}[H]
\centering
\resizebox{\textwidth}{!}{%
\begin{tikzpicture}[
  font=\sffamily\scriptsize,
  role/.style={draw=bcNavy,fill=bcNavy,text=white,rounded corners=2pt,
               text width=26mm,align=center,minimum height=7mm,
               inner sep=2pt,font=\sffamily\tiny\bfseries},
  stage/.style={draw=bcBlue,fill=bcSand,rounded corners=2.5pt,text width=26mm,
                align=center,minimum height=15mm,inner sep=2.5pt,font=\sffamily\tiny},
  dec/.style={draw=bcAmber,fill=bcAmber!18,rounded corners=2pt,text width=26mm,
              align=center,minimum height=11mm,inner sep=2pt,font=\sffamily\tiny},
  exc/.style={draw=bcRed,fill=bcRed!10,rounded corners=2pt,text width=30mm,
              align=center,minimum height=7mm,inner sep=2pt,font=\sffamily\tiny},
  ttl/.style={font=\sffamily\tiny\bfseries,text=bcNavy,anchor=west},
  fl/.style={draw=bcBlue,-{Latex[length=1.6mm]},line width=0.5pt},
  fle/.style={draw=bcRed,-{Latex[length=1.6mm]},line width=0.5pt,dashed}
]
\node[ttl] at (-8.6,3.5) {HUMAN ROLE};
\node[role] (r1) at (-6.4,2.75) {Lab technician};
\node[role] (r2) at (-3.2,2.75) {Bioinformatician};
\node[role] (r3) at (0.0,2.75)  {Bioinformatician};
\node[role] (r4) at (3.2,2.75)  {Variant analyst};
\node[role] (r5) at (6.4,2.75)  {Domain scientist};

\node[ttl] at (-8.6,1.85) {PIPELINE STAGE};
\node[stage] (s1) at (-6.4,0.85)
  {\textbf{QC \& preprocessing}\\FastQC / MultiQC review;\\adapter \& quality trimming};
\node[stage] (s2) at (-3.2,0.85)
  {\textbf{Alignment}\\reference selection;\\aligner parameters;\\duplicate marking};
\node[stage] (s3) at (0.0,0.85)
  {\textbf{Variant calling}\\caller selection;\\calling parameters;\\hard / soft filtering};
\node[stage] (s4) at (3.2,0.85)
  {\textbf{Annotation}\\functional annotation;\\frequency filtering;\\effect prioritisation};
\node[stage] (s5) at (6.4,0.85)
  {\textbf{Reporting}\\result assembly;\\interpretation;\\downstream handoff};

\foreach \a/\b in {s1/s2, s2/s3, s3/s4, s4/s5}{\draw[fl] (\a.east) -- (\b.west);}
\foreach \r/\s in {r1/s1, r2/s2, r3/s3, r4/s4, r5/s5}{\draw[fl,draw=bcGrey!60] (\r.south) -- (\s.north);}

\node[ttl] at (-8.6,-0.55) {MANUAL DECISION};
\node[dec] (d1) at (-6.4,-1.5) {which thresholds?\\resequence or proceed?};
\node[dec] (d2) at (-3.2,-1.5) {which reference?\\accept mapping rate?};
\node[dec] (d3) at (0.0,-1.5)  {which filters?\\trust marginal calls?};
\node[dec] (d4) at (3.2,-1.5)  {which variants\\warrant review?};
\node[dec] (d5) at (6.4,-1.5)  {what is reportable?\\what is caveated?};
\foreach \s/\d in {s1/d1, s2/d2, s3/d3, s4/d4, s5/d5}{\draw[fl,draw=bcAmber] (\s.south) -- (\d.north);}

\node[anchor=center,font=\sffamily\tiny\itshape,text=bcGrey] at (-4.8,-2.35) {manual handoff};
\node[anchor=center,font=\sffamily\tiny\itshape,text=bcGrey] at (-1.6,-2.35) {manual handoff};
\node[anchor=center,font=\sffamily\tiny\itshape,text=bcGrey] at (1.6,-2.35)  {manual handoff};
\node[anchor=center,font=\sffamily\tiny\itshape,text=bcGrey] at (4.8,-2.35)  {manual handoff};
\draw[bcGrey!45,dashed] (-8.0,-2.75) -- (8.0,-2.75);

\node[ttl] at (-8.6,-3.15) {EXCEPTION HANDLING --- reactive, detected after the fact};
\node[exc] (e1) at (-5.6,-3.95) {low-coverage samples};
\node[exc] (e2) at (-1.9,-3.95) {contamination / index hopping};
\node[exc] (e3) at (1.8,-3.95)  {aberrant call-rate shifts};
\node[exc] (e4) at (5.5,-3.95)  {batch effects across runs};
\foreach \e in {e1,e2,e3,e4}{\draw[fle] (\e.north) -- ++(0,0.55);}
\node[anchor=west,font=\sffamily\tiny\itshape,text=bcRed,text width=150mm] at (-8.6,-4.7)
  {Exceptions surface only when an operator notices an anomaly in a report or
   when a downstream result appears implausible; there is no systematic
   mechanism for proactive detection across the run.};
\end{tikzpicture}}
\caption{Manual DNA sequencing pipeline workflow. Laboratory technicians,
bioinformaticians, variant analysts, and domain scientists execute sequential
stages --- quality control, alignment, variant calling, annotation, and
reporting --- with tool execution largely automated by workflow
managers~\cite{nextflow, snakemake, nfcore} but the surrounding decision layer
(amber) remaining manual and operator-dependent. Handoffs between stages are
informal, and the reasoning behind each decision is rarely recorded in a form
that survives into the next stage. Exceptions such as low coverage,
contamination, aberrant call rates, and batch effects are detected reactively
--- typically after they have already propagated downstream --- limiting the
ability to intervene before results are affected.}
\label{fig:manualflow}
\end{figure}

It is important to be precise about what is and is not manual here. Tool
\textit{execution} is largely automated: workflow managers such as Nextflow
and Snakemake~\cite{nextflow, snakemake}, and curated pipeline collections
such as nf-core~\cite{nfcore}, reliably orchestrate the invocation of FastQC,
BWA, GATK, and their counterparts across large sample batches. What remains
manual is the \textit{decision layer} surrounding that execution: selecting
thresholds appropriate to a given sample and platform, interpreting quality
reports, adjudicating borderline variant calls, diagnosing pipeline failures,
and determining which results warrant expert attention. As depicted in
Figure~\ref{fig:manualflow}, these decisions are made sequentially by distinct
human roles --- from laboratory technicians assessing run quality through to
analysts reviewing annotated variant sets --- with no automated reasoning
layer connecting them. The decisions are highly repetitive, require synthesis
of information across heterogeneous outputs, and are sensitive to operator
experience in ways that manual processes cannot standardise at scale. These
characteristics --- continuous monitoring requirements, repeated
judgment-intensive decisions, cross-source reasoning, and the need for human
oversight at critical points --- are precisely the conditions under which
agentic AI workflows deliver the greatest operational value, as identified in
our earlier work on agentic AI transition
frameworks~\cite{agentic-ai-transition-organization,
agentic-workflow-practicle-guide}. The manual processes underlying each stage
are described in detail in the following subsections.

\subsection{Raw Read Quality Control and Preprocessing}
\label{subsec:manual_qc}

The pipeline begins when sequencer output is demultiplexed and quality
assessed. Bioinformaticians run FastQC on each sample and aggregate the
results using MultiQC~\cite{fastqc, multiqc}, producing reports covering
per-base quality distributions, GC content, adapter contamination, duplication
levels, and overrepresented sequences. Interpreting these reports is a manual
task requiring contextual judgment: a GC distribution that would indicate
contamination in a whole-genome sample may be entirely expected in an
amplicon or reduced-representation library, and a duplication rate that is
alarming in one protocol is normal in another. Trimming decisions --- which
adapter sequences to remove, what quality threshold to apply, what minimum
read length to retain --- are typically made by applying a laboratory default
and adjusting when results appear anomalous~\cite{trimmomatic, fastp}.

The consequences of misjudgment at this stage are asymmetric and easily
overlooked. Over-aggressive trimming discards genuine signal and reduces
effective coverage, while under-trimming propagates adapter and low-quality
sequence into alignment, inflating mismatch rates and depressing mapping
quality in ways that are subsequently attributed to sample quality rather than
to preprocessing. In high-throughput settings, per-sample review is often
infeasible, and samples are processed in batches under a single parameter set
that suits the majority but not the outliers. Which samples are genuine
outliers requiring individual attention, as opposed to acceptable variation,
is a determination made by eye from aggregate plots.

\subsection{Reference Alignment and Parameter Selection}
\label{subsec:manual_align}

Once reads are preprocessed, they are aligned to a reference genome using
tools such as BWA-MEM or Bowtie2~\cite{bwa, bowtie2}, followed by sorting,
duplicate marking, and, in some workflows, base quality score
recalibration~\cite{samtools, gatk}. Reference selection is itself a
consequential manual decision. In clinical and model-organism contexts a
standard reference assembly is available, but in agricultural and
non-model-organism genomics --- including crop and horticultural breeding
programmes --- the appropriate reference may be an assembly of a related
cultivar or species, and the choice materially affects mapping rate and
downstream variant discovery, particularly in divergent regions.

Following alignment, the bioinformatician reviews mapping statistics:
alignment rate, properly paired proportion, mean and distribution of coverage,
insert size distribution, and duplication rate. Deciding whether these
statistics are acceptable requires joint reasoning across several of them at
once --- a reduced mapping rate accompanied by an unusual insert size
distribution suggests a library preparation issue, whereas the same mapping
rate with a normal insert distribution and elevated unmapped reads may
indicate contamination or reference mismatch. This joint interpretation is
performed manually, is rarely documented, and is frequently deferred under
throughput pressure with the sample simply passed forward.

\subsection{Variant Calling and Filtering}
\label{subsec:manual_calling}

Variant calling converts aligned reads into a set of putative genetic
variants, using callers such as GATK HaplotypeCaller, DeepVariant, or
bcftools~\cite{gatk, deepvariant, samtools}. Caller selection depends on
sequencing depth, platform, ploidy, and whether germline or somatic variants
are sought, and the choice is typically fixed by laboratory convention rather
than reasoned per experiment. Calling parameters --- minimum base and mapping
quality, ploidy assumptions, and handling of multi-allelic sites --- are
similarly inherited from prior runs.

Filtering is the most judgment-intensive step in the pipeline. Raw call sets
contain substantial numbers of false positives, and filtering strategies range
from hard threshold filters on annotations such as depth, quality by depth,
strand bias, and mapping quality, to model-based approaches such as variant
quality score recalibration where sufficient training data
exists~\cite{gatk, vcfspec}. Selecting thresholds involves an explicit
trade-off between sensitivity and specificity whose appropriate balance
depends on the downstream application --- a screening context tolerates false
positives that a reporting context does not --- and this trade-off is
typically resolved by convention rather than by per-run reasoning.

Two properties make this stage particularly consequential. Filtering decisions
are \textit{silent}: a filtered variant simply does not appear downstream, and
its absence leaves no trace unless deliberately recorded. And they are
\textit{cumulative}: every subsequent analysis inherits the filtered set
without visibility into what was removed or why. In manual workflows, the
rationale for a given threshold is frequently held only in the operator's
working memory or an informal note, and is not recoverable when a result is
later questioned.

\subsection{Annotation and Functional Interpretation}
\label{subsec:manual_annotation}

Filtered variants are annotated with functional consequence predictions,
population allele frequencies, and, where relevant, prior evidence from
curated databases, using tools such as the Ensembl Variant Effect Predictor,
SnpEff, or ANNOVAR against resources including gnomAD and dbSNP~\cite{vep,
snpeff, annovar, gnomad, dbsnp}. Annotation execution is well automated; the
manual work lies in the prioritisation that follows.

An annotated call set for a single sample commonly contains many thousands of
variants, of which a small number are of interest for the study at hand.
Reducing this set requires layered filtering on consequence severity,
population frequency, region of interest, inheritance pattern, or association
with traits under investigation. Analysts construct these filters manually,
frequently in spreadsheets or ad hoc scripts, and the specific combination
applied varies with the analyst and the question. In trait-mapping and
breeding contexts --- linking genotype to phenotypes such as yield, fruit
quality, or disease resistance --- an additional manual step joins the variant
set to phenotypic records held in separate systems, a cross-source integration
that is laborious and error-prone. The filtering logic is rarely captured
alongside the results, so a prioritised variant list is often difficult to
reproduce even by its author.

\subsection{Pipeline Monitoring and Exception Handling}
\label{subsec:manual_exception}

Throughout the pipeline, exceptions arise that require human diagnosis:
failed or stalled jobs, samples with anomalously low coverage, contamination
or index hopping between multiplexed samples, unexpected shifts in variant
call rate, and batch effects distinguishing one sequencing run from
another~\cite{multiqc, nfcore}. Workflow managers detect and report
\textit{execution} failures reliably --- a job that crashes is surfaced
immediately --- but \textit{analytical} anomalies, where every job completes
successfully yet the results are wrong, are not detected by execution
monitoring at all.

In manual workflows, detection of such anomalies depends on an operator
noticing an irregularity in an aggregate report, or on a downstream analysis
producing an implausible result that prompts retrospective investigation.
There is no systematic mechanism for proactive detection across the full
pipeline, which means that many issues are identified only after affected
results have already been used. Once detected, diagnosis requires tracing
back through pipeline stages to determine whether the cause is sample quality,
a preprocessing decision, an alignment problem, or a calling parameter --- a
reconstruction that is time-consuming and depends on records that manual
workflows frequently do not retain. The reactive and reconstruction-dependent
nature of manual exception handling limits the ability to intervene before
compromised results propagate downstream.

\section{Delegating Manual Sequencing Processes to AI Agents}
\label{sec:framework}

Once the manual workflows are well understood, the next step in the agentic AI
transition is to systematically delegate them to a coordinated network of
autonomous AI agents~\cite{agent-survey, agentic-ai-transition-organization}.
This delegation goes beyond automating isolated tasks --- it requires
decomposing human-performed decision-making into distinct reasoning steps,
decision points, and coordination actions that can be meaningfully assigned to
specialised agents, each operating within a clearly defined
scope~\cite{sapkota2025, agentic-ai-workflow-patterns}. The goal is not to
replicate the existing manual process in software, but to capture the
underlying intent and judgment that guide human work across the sequencing
lifecycle.

One design decision governs the entire framework and warrants statement before
the agents are described. \textbf{BaseCamp agents do not perform sequence
analysis.} Alignment is performed by BWA, variant calling by GATK or
DeepVariant, annotation by VEP --- established, independently validated tools
whose correctness properties are well characterised and whose outputs are
reproducible~\cite{bwa, gatk, deepvariant, vep}. The agents select among these
tools, configure them, interpret their output, decide what follows, and
determine what warrants human attention. This division is deliberate and
load-bearing: it confines LLM reasoning to the judgment layer where it is
reliable, preserves the reproducibility guarantees the existing tooling already
provides, and ensures that a BaseCamp result is a result an established tool
produced rather than one a language model generated.

In practice, delegation begins by identifying the cognitive responsibilities
embedded within each stage of the manual workflow --- quality interpretation,
alignment assessment, calling and filtering strategy, variant prioritisation,
anomaly detection, and result communication. Each responsibility is mapped to
a dedicated agent with a well-defined role, a set of inputs and outputs, and
access via MCP to the tools and data sources required to execute it. Agents
operate in coordination, passing structured intermediate outputs between
stages so that downstream agents reason with full upstream context rather than
in isolation --- a property manual handoffs conspicuously lack, as noted in
Section~\ref{sec:manual}.

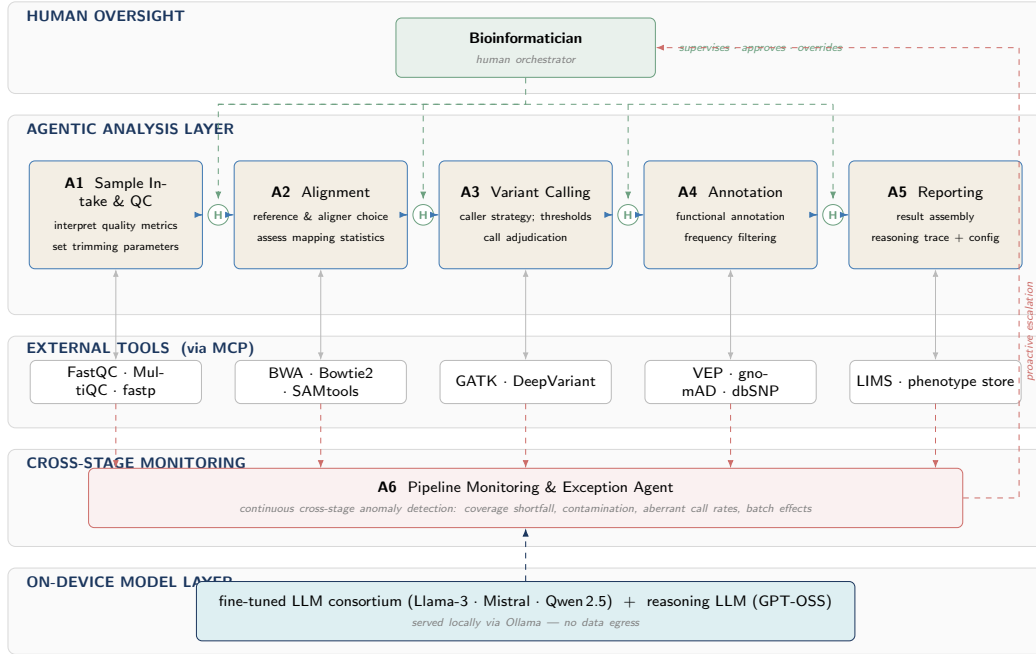
\begin{figure*}[!htb]
\centering
\resizebox{\textwidth}{!}{%
\begin{tikzpicture}[
  font=\sffamily\scriptsize,
  ag/.style={draw=bcBlue,fill=bcSand,rounded corners=3pt,text width=30mm,
             align=center,minimum height=20mm,inner sep=3pt,font=\sffamily\scriptsize},
  mon/.style={draw=bcRed!75,fill=bcRed!7,rounded corners=3pt,text width=160mm,
              align=center,minimum height=11mm,inner sep=3pt,font=\sffamily\scriptsize},
  tool/.style={draw=bcGrey!55,fill=white,rounded corners=3pt,text width=30mm,
               align=center,minimum height=8mm,inner sep=2.5pt,font=\sffamily\scriptsize},
  mdl/.style={draw=bcNavy,fill=bcLight!40,rounded corners=3pt,text width=120mm,
              align=center,minimum height=11mm,inner sep=3pt,font=\sffamily\scriptsize},
  gate/.style={draw=bcGreen!80,fill=white,circle,inner sep=1.6pt,
               font=\sffamily\tiny\bfseries,text=bcGreen!80},
  hum/.style={draw=bcGreen!80,fill=bcGreen!12,rounded corners=3pt,text width=46mm,
              align=center,minimum height=11mm,inner sep=3pt,font=\sffamily\scriptsize},
  band/.style={rounded corners=4pt,draw=bcGrey!30,fill=bcGrey!4,line width=0.4pt},
  hdr/.style={font=\sffamily\scriptsize\bfseries,text=bcNavy,anchor=west},
  note/.style={font=\sffamily\tiny\itshape,text=bcGrey},
  fl/.style={draw=bcBlue,-{Latex[length=1.6mm]},line width=0.55pt},
  flt/.style={draw=bcGrey!55,{Latex[length=1.4mm]}-{Latex[length=1.4mm]},line width=0.45pt},
  flg/.style={draw=bcGreen!80,-{Latex[length=1.5mm]},line width=0.5pt,dashed},
  flr/.style={draw=bcRed!75,-{Latex[length=1.5mm]},line width=0.5pt,dashed}
]

\draw[band] (-9.6,5.05) rectangle (9.6,3.35);    
\draw[band] (-9.6,2.95) rectangle (9.6,-0.75);   
\draw[band] (-9.6,-1.15) rectangle (9.6,-2.85);  
\draw[band] (-9.6,-3.25) rectangle (9.6,-5.05);  
\draw[band] (-9.6,-5.45) rectangle (9.6,-7.05);  

\node[hdr] at (-9.4,4.8)  {HUMAN OVERSIGHT};
\node[hdr] at (-9.4,2.7)  {AGENTIC ANALYSIS LAYER};
\node[hdr] at (-9.4,-1.4) {EXTERNAL TOOLS \ (via MCP)};
\node[hdr] at (-9.4,-3.5) {CROSS-STAGE MONITORING};
\node[hdr] at (-9.4,-5.7) {ON-DEVICE MODEL LAYER};

\node[hum] (h) at (0,4.2) {\textbf{Bioinformatician}\\[1pt]{\tiny\itshape\color{bcGrey} human orchestrator}};
\node[note,text=bcGreen!80,anchor=west] at (2.7,4.2) {supervises $\cdot$ approves $\cdot$ overrides};

\node[ag] (a1) at (-7.6,1.1) {\textbf{A1} \ Sample Intake \& QC\\[3pt]
  {\tiny interpret quality metrics}\\[1.5pt]{\tiny set trimming parameters}};
\node[ag] (a2) at (-3.8,1.1) {\textbf{A2} \ Alignment\\[3pt]
  {\tiny reference \& aligner choice}\\[1.5pt]{\tiny assess mapping statistics}};
\node[ag] (a3) at ( 0.0,1.1) {\textbf{A3} \ Variant Calling\\[3pt]
  {\tiny caller strategy; thresholds}\\[1.5pt]{\tiny call adjudication}};
\node[ag] (a4) at ( 3.8,1.1) {\textbf{A4} \ Annotation\\[3pt]
  {\tiny functional annotation}\\[1.5pt]{\tiny frequency filtering}};
\node[ag] (a5) at ( 7.6,1.1) {\textbf{A5} \ Reporting\\[3pt]
  {\tiny result assembly}\\[1.5pt]{\tiny reasoning trace + config}};

\foreach \a/\b/\x in {a1/a2/-5.7, a2/a3/-1.9, a3/a4/1.9, a4/a5/5.7}{%
  \draw[fl] (\a.east) -- (\x-0.28,1.1);
  \node[gate] (g\x) at (\x,1.1) {H};
  \draw[fl] (\x+0.28,1.1) -- (\b.west);}

\draw[flg] (h.south) -- (0,3.15) -- (-5.7,3.15) -- (-5.7,1.38);
\draw[flg] (-1.9,3.15) -- (-1.9,1.38);
\draw[flg] ( 1.9,3.15) -- ( 1.9,1.38);
\draw[flg] ( 5.7,3.15) -- ( 5.7,1.38);
\draw[draw=bcGreen!80,dashed,line width=0.5pt] (-5.7,3.15) -- (5.7,3.15);

\node[tool] (t1) at (-7.6,-2.0) {FastQC $\cdot$ MultiQC $\cdot$ fastp};
\node[tool] (t2) at (-3.8,-2.0) {BWA $\cdot$ Bowtie2 $\cdot$ SAMtools};
\node[tool] (t3) at ( 0.0,-2.0) {GATK $\cdot$ DeepVariant};
\node[tool] (t4) at ( 3.8,-2.0) {VEP $\cdot$ gnomAD $\cdot$ dbSNP};
\node[tool] (t5) at ( 7.6,-2.0) {LIMS $\cdot$ phenotype store};
\foreach \a/\t in {a1/t1, a2/t2, a3/t3, a4/t4, a5/t5}{\draw[flt] (\a.south) -- (\t.north);}

\node[mon] (a6) at (0,-4.15) {\textbf{A6} \ Pipeline Monitoring \& Exception Agent\\[1pt]
  {\tiny\itshape\color{bcGrey} continuous cross-stage anomaly detection: coverage shortfall,
   contamination, aberrant call rates, batch effects}};

\foreach \x in {-7.6,-3.8,0.0,3.8,7.6}{\draw[flr] (\x,-2.4) -- (\x,-3.6);}

\draw[flr] (a6.east) -- (9.15,-4.15) -- (9.15,4.2) -- (h.east);
\node[note,text=bcRed!75,anchor=east,rotate=90] at (9.35,0.0) {proactive escalation};

\node[mdl] (ml) at (0,-6.25) {fine-tuned LLM consortium
  (Llama-3 $\cdot$ Mistral $\cdot$ Qwen\,2.5) \ + \ reasoning LLM (GPT-OSS)\\[1pt]
  {\tiny\itshape\color{bcGrey} served locally via Ollama --- no data egress}};
\draw[fl,draw=bcNavy,dashed] (ml.north) -- (a6.south);
\end{tikzpicture}}
\caption{The BaseCamp agentic AI workflow for automated DNA sequencing
pipelines. Six specialised agents replace the manual decision layer of
Figure~\ref{fig:manualflow}: five operate sequentially across the pipeline
lifecycle, passing structured outputs so that each reasons with full upstream
context, while the Pipeline Monitoring and Exception Agent observes all stages
continuously and escalates anomalies proactively rather than awaiting
retrospective discovery. Agents invoke established bioinformatics
tools~\cite{fastqc, bwa, gatk, deepvariant, vep} through MCP
servers~\cite{mcp1, mcp2, mcc} rather than performing analysis themselves.
Human approval gates (H) are embedded at consequential transitions, and all
agent reasoning is served by a locally deployed fine-tuned LLM consortium
coordinated by a reasoning LLM.}
\label{fig:agentflow}
\end{figure*}

Figure~\ref{fig:agentflow} illustrates how the manual pipeline of
Figure~\ref{fig:manualflow} is decomposed into the BaseCamp agentic workflow.
Six specialised agents --- a Sample Intake and QC Agent, an Alignment Agent, a
Variant Calling Agent, an Annotation and Interpretation Agent, a Pipeline
Monitoring and Exception Agent, and a Reporting Agent --- operate as a
coordinated system under human supervision, each connected to relevant tools
and data sources through MCP servers~\cite{mcp1, mcp2, mcc}.

\subsection{Sample Intake and QC Agent}
\label{subsec:a1}

The Sample Intake and QC Agent is responsible for ingesting demultiplexed
sequencer output, invoking quality assessment tooling, and reasoning over the
resulting metrics to determine preprocessing parameters and sample
disposition. The agent interfaces with FastQC, MultiQC, and preprocessing
tools via MCP server connections~\cite{fastqc, multiqc, fastp}, retrieving
per-base quality distributions, GC content, adapter contamination, duplication
levels, and overrepresented sequence reports.

Its principal contribution is contextual interpretation. As described in
Section~\ref{subsec:manual_qc}, a metric that indicates a problem under one
library preparation is expected under another, and this distinction is exactly
what a fixed threshold cannot encode. Using a fine-tuned LLM trained on
historical QC reports paired with the dispositions bioinformaticians assigned
them, the agent reasons jointly over the metric profile, the declared library
type, the sequencing platform, and the sample's position within the batch, and
produces a structured determination: recommended trimming parameters with
explicit justification, a pass, review, or fail disposition, and an
identification of samples whose profile deviates from the batch in ways
warranting individual attention. Because per-sample review is infeasible
manually at scale, the agent's ability to reason per sample rather than per
batch is where the operational gain arises. Outputs pass to the Alignment
Agent; samples flagged as failing or ambiguous are escalated for human
adjudication.

\subsection{Alignment Agent}
\label{subsec:a2}

The Alignment Agent selects an appropriate reference and aligner
configuration, invokes alignment and post-processing, and assesses the
resulting statistics. It interfaces with BWA, Bowtie2, and SAMtools via MCP
connections~\cite{bwa, bowtie2, samtools}, and with reference genome
repositories and, where applicable, laboratory information management systems
for sample provenance.

Two reasoning tasks distinguish this agent. The first is reference selection,
which as noted in Section~\ref{subsec:manual_align} is consequential in
non-model-organism contexts where the appropriate reference may be an assembly
of a related cultivar or species; the agent reasons over declared sample
provenance, available assemblies, and observed mapping behaviour to recommend
a reference with explicit justification. The second is joint interpretation of
alignment statistics. Rather than comparing each statistic against an
independent threshold, the agent reasons across mapping rate, properly paired
proportion, insert size distribution, coverage uniformity, and duplication
rate simultaneously, and distinguishes characteristic failure signatures ---
a library preparation problem, contamination, or reference divergence ---
that individual thresholds cannot separate. This joint reasoning is precisely
what manual review performs inconsistently under throughput pressure. Aligned
outputs and the accompanying assessment pass to the Variant Calling Agent.

\subsection{Variant Calling Agent}
\label{subsec:a3}

The Variant Calling Agent determines calling strategy, invokes the selected
caller, and reasons over filtering. It interfaces with GATK, DeepVariant, and
bcftools via MCP connections~\cite{gatk, deepvariant, samtools}, receiving
aligned data and quality context from upstream agents.

The agent reasons over caller selection given depth, platform, ploidy, and
whether germline or somatic variants are sought --- a determination typically
fixed by convention rather than reasoned per experiment in manual workflows
--- and then addresses filtering, the pipeline's most judgment-intensive step.
A fine-tuned LLM enables the agent to reason about the sensitivity--specificity
trade-off in light of the stated downstream application, recommending
thresholds on depth, quality by depth, strand bias, and mapping quality with
explicit justification for each.

Two design requirements follow from the properties identified in
Section~\ref{subsec:manual_calling}. Because filtering decisions are silent,
the agent emits an explicit \textit{filtering ledger} recording every
threshold applied, the number of calls it removed, and the rationale --- so
that what was excluded remains inspectable rather than merely absent. And
because they are cumulative, the ledger accompanies the call set downstream
rather than remaining with the operator. Marginal calls falling near a
threshold boundary are flagged individually for human adjudication rather than
silently resolved. The filtered call set, the ledger, and the flagged calls
pass to the Annotation and Interpretation Agent.

\subsection{Annotation and Interpretation Agent}
\label{subsec:a4}

The Annotation and Interpretation Agent enriches the filtered call set with
functional consequence predictions, population allele frequencies, and prior
database evidence, and reasons over the enriched set to produce a prioritised
shortlist. It interfaces with VEP, SnpEff, and ANNOVAR, and with gnomAD and
dbSNP, via MCP connections~\cite{vep, snpeff, annovar, gnomad, dbsnp}.

Annotation execution is well automated already; the agent's contribution is
the prioritisation that follows. Reasoning over consequence severity,
population frequency, region of interest, inheritance pattern, and the
study's stated objective, it reduces a call set of many thousands to a
reviewable shortlist, recording the filter logic applied at each step so that
the shortlist is reproducible --- addressing the reproducibility gap noted in
Section~\ref{subsec:manual_annotation}, where manually constructed
prioritisations are frequently difficult to reproduce even by their author. In
trait-mapping and breeding contexts the agent additionally joins the variant
set to phenotypic records retrieved from separate systems, performing the
genotype--phenotype integration that is otherwise a laborious manual
cross-source step. The prioritised set and its filter provenance pass to the
Reporting Agent.

\subsection{Pipeline Monitoring and Exception Agent}
\label{subsec:a5}

The Pipeline Monitoring and Exception Agent observes all stages continuously,
detecting conditions that fall outside expected operating parameters and
require human judgment. These include coverage shortfalls that will compromise
downstream calling, contamination or index hopping between multiplexed
samples, unexpected shifts in variant call rate relative to comparable runs,
batch effects distinguishing one sequencing run from another, and stalled or
failed pipeline stages.

Its role is defined by a gap that workflow managers do not address. As noted
in Section~\ref{subsec:manual_exception}, execution monitoring reliably
surfaces jobs that crash, but \textit{analytical} anomalies --- where every
job completes successfully and the results are nonetheless wrong --- are
invisible to it. The agent reasons across stages rather than within them,
comparing observed metrics against expectations derived from the run's own
distribution and from historical comparable runs, and detecting inconsistency
patterns that no single stage would flag in isolation. Rather than awaiting
retrospective discovery, it generates structured, human-readable alerts
prioritised by operational urgency, each citing the specific metric and
comparison that triggered it, and each proposing candidate diagnoses ---
sample quality, a preprocessing decision, an alignment problem, or a calling
parameter --- with the supporting evidence for each. This proposed diagnosis
substantially reduces the backward-tracing burden that manual exception
handling imposes.

\subsection{Reporting Agent}
\label{subsec:a6}

The Reporting Agent assembles the final artifact: the prioritised variant set,
the per-stage reasoning traces accumulated across the pipeline, the filtering
ledger, outstanding flagged items awaiting adjudication, and the complete
configuration under which the run executed --- tool versions and parameters,
model versions, and adapter revisions.

The configuration record is not incidental. Where a sequencing result informs
a consequential downstream decision and is subsequently questioned, the
relevant question is not only what the pipeline produced but under what
configuration, and whether a different defensible configuration would have
produced something else. Recording the settings makes that question
answerable, which is a precondition for reproducible scientific
use~\cite{nextflow, snakemake, nfcore}. The agent generates reports targeted
to their audience --- a technical run summary for the bioinformatics team, a
result-focused summary for the requesting scientist --- with each reported
finding traceable to the pipeline stage and decision that produced it.

\subsection{Agent Decomposition and Workflow Coordination}
\label{subsec:coordination}

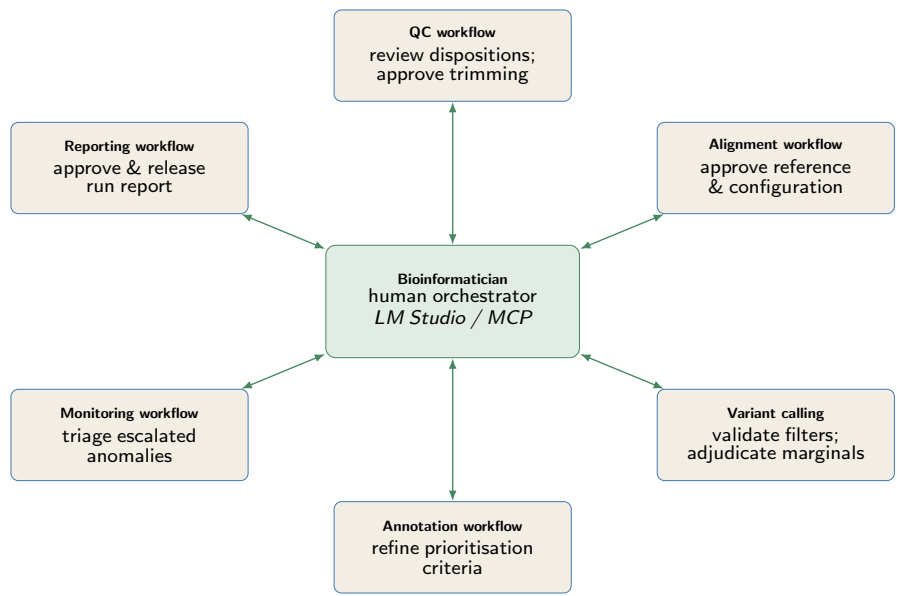
\begin{figure}[H]
\centering
\resizebox{0.9\textwidth}{!}{%
\begin{tikzpicture}[
  font=\sffamily\scriptsize,
  hum/.style={draw=bcGreen,fill=bcGreen!16,rounded corners=4pt,text width=34mm,
              align=center,minimum height=16mm,inner sep=3pt,font=\sffamily\tiny\bfseries},
  wf/.style={draw=bcBlue,fill=bcSand,rounded corners=3pt,text width=32mm,
             align=center,minimum height=13mm,inner sep=2.5pt,font=\sffamily\tiny},
  fl/.style={draw=bcGreen,{Latex[length=1.5mm]}-{Latex[length=1.5mm]},line width=0.55pt}
]
\node[hum] (h) at (0,0) {Bioinformatician\\{\scriptsize\mdseries human orchestrator}\\[2pt]
  {\scriptsize\mdseries\itshape LM Studio / MCP}};

\node[wf] (w1) at (0,3.5)      {\textbf{QC workflow}\\{\scriptsize review dispositions;\\approve trimming}};
\node[wf] (w2) at (4.6,1.9)    {\textbf{Alignment workflow}\\{\scriptsize approve reference\\\& configuration}};
\node[wf] (w3) at (4.6,-1.9)   {\textbf{Variant calling}\\{\scriptsize validate filters;\\adjudicate marginals}};
\node[wf] (w4) at (0,-3.5)     {\textbf{Annotation workflow}\\{\scriptsize refine prioritisation\\criteria}};
\node[wf] (w5) at (-4.6,-1.9)  {\textbf{Monitoring workflow}\\{\scriptsize triage escalated\\anomalies}};
\node[wf] (w6) at (-4.6,1.9)   {\textbf{Reporting workflow}\\{\scriptsize approve \& release\\run report}};

\foreach \w in {w1,w2,w3,w4,w5,w6}{\draw[fl] (h) -- (\w);}

\node[anchor=north,font=\sffamily\tiny\itshape,text=bcGrey,text width=130mm,align=center]
  at (0,-4.6)
  {Each workflow is exposed as an independent MCP server, so a disputed stage
   can be interrogated or re-run without recomputing the pipeline.};
\end{tikzpicture}}
\caption{A single bioinformatician orchestrating multiple specialised BaseCamp
agentic workflows. Each workflow automates a distinct pipeline function ---
quality control, alignment, variant calling, annotation, monitoring, and
reporting --- and is exposed through an independent MCP server~\cite{mcp1,
mcp2, mcc} for human-supervised coordination, approval, and intervention via a
unified natural language interface in LM Studio~\cite{lm-studio}. The human
role shifts from executing each stage to supervising all of them, with
authority retained at every consequential decision point.}
\label{fig:orchestrator}
\end{figure}

The agent-based decomposition introduces modularity, parallelism, and
controllability into the sequencing pipeline. Rather than requiring a human to
coordinate between stages, each agent is exposed as an independent workflow
accessible through an MCP server, enabling a single bioinformatician to
orchestrate, supervise, and intervene across all six agents via a unified
natural language interface, as illustrated in
Figure~\ref{fig:orchestrator}. The orchestrator interacts at defined
checkpoints --- reviewing QC dispositions, approving alignment
configurations, validating filtering thresholds, adjudicating flagged
variants, and responding to escalated exceptions --- without participating in
the routine execution of any stage~\cite{agentic-workflow-practicle-guide}.

Because servers are individually addressable, a bioinformatician who disputes
one stage can re-run that stage alone rather than recomputing the entire
pipeline --- the common case when a run is questioned in part rather than in
whole, and a substantial practical advantage in a domain where full pipeline
re-execution is measured in hours or days. Individual agents can likewise be
refined, retrained, or extended independently: a variant caller can be
upgraded, or a filtering policy revised, without redesigning the workflow.
Agents share state through structured intermediate outputs, so downstream
agents reason with full upstream context --- the QC Agent's assessment of
library quality informs the Alignment Agent's interpretation of mapping rate,
which in turn informs the Variant Calling Agent's confidence in marginal
calls. This contextual continuity is what manual handoffs, which transmit
results without the reasoning that produced them, systematically lose.

\subsection{Responsible and Explainable AI Agents}
\label{subsec:responsibleagents}

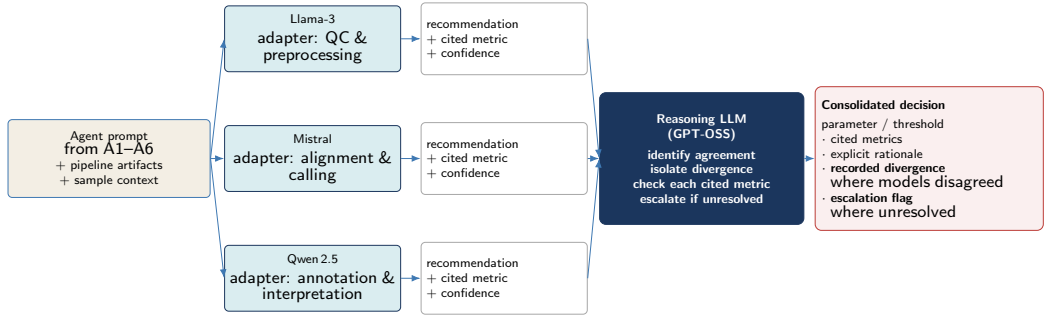
\begin{figure}[H]
\centering
\resizebox{\textwidth}{!}{%
\begin{tikzpicture}[
  font=\sffamily\scriptsize,
  pr/.style={draw=bcBlue,fill=bcSand,rounded corners=2pt,text width=32mm,
             align=center,minimum height=13mm,inner sep=2.5pt,font=\sffamily\tiny},
  mdl/.style={draw=bcNavy,fill=bcLight!45,rounded corners=2pt,text width=28mm,
              align=center,minimum height=11mm,inner sep=2pt,font=\sffamily\tiny},
  res/.style={draw=bcGrey!70,fill=white,rounded corners=2pt,text width=26mm,
              align=left,minimum height=12mm,inner sep=2.5pt,font=\sffamily\tiny},
  rsn/.style={draw=bcNavy,fill=bcNavy,text=white,rounded corners=3pt,text width=32mm,
              align=center,minimum height=22mm,inner sep=3pt,font=\sffamily\tiny\bfseries},
  fin/.style={draw=bcRed,fill=bcRed!8,rounded corners=3pt,text width=36mm,
              align=left,minimum height=24mm,inner sep=3pt,font=\sffamily\tiny},
  fl/.style={draw=bcBlue,-{Latex[length=1.5mm]},line width=0.45pt}
]
\node[pr] (p) at (-6.7,0)
  {Agent prompt\\{\scriptsize from A1--A6}\\+ pipeline artifacts\\+ sample context};

\node[mdl] (m1) at (-3.3,2.0)  {Llama-3\\{\scriptsize adapter: QC \&\\preprocessing}};
\node[mdl] (m2) at (-3.3,0)    {Mistral\\{\scriptsize adapter: alignment \&\\calling}};
\node[mdl] (m3) at (-3.3,-2.0) {Qwen\,2.5\\{\scriptsize adapter: annotation \&\\interpretation}};

\node[res] (o1) at (-0.1,2.0)  {recommendation\\+ cited metric\\+ confidence};
\node[res] (o2) at (-0.1,0)    {recommendation\\+ cited metric\\+ confidence};
\node[res] (o3) at (-0.1,-2.0) {recommendation\\+ cited metric\\+ confidence};

\node[rsn] (r) at (3.2,0)
  {\textbf{Reasoning LLM}\\(GPT-OSS)\\[3pt]
   identify agreement\\
   isolate divergence\\
   check each cited metric\\
   \textit{escalate if unresolved}};

\node[fin] (f) at (7.0,0)
  {\textbf{Consolidated decision}\\[2pt]
   parameter / threshold\\
   $\cdot$ cited metrics\\
   $\cdot$ explicit rationale\\
   $\cdot$ \textbf{recorded divergence}\\
   \ \ {\scriptsize where models disagreed}\\
   $\cdot$ \textbf{escalation flag}\\
   \ \ {\scriptsize where unresolved}};

\foreach \m in {m1,m2,m3}{\draw[fl] (p.east) -- (\m.west);}
\foreach \a/\b in {m1/o1, m2/o2, m3/o3}{\draw[fl] (\a.east) -- (\b.west);}
\foreach \o in {o1,o2,o3}{\draw[fl] (\o.east) -- (r.west);}
\draw[fl] (r.east) -- (f.west);

\node[anchor=north,font=\sffamily\tiny\itshape,text=bcGrey,text width=150mm,align=center]
  at (0,-3.3)
  {Unresolved divergence between models is surfaced to the human orchestrator
   rather than averaged away: a decision the models cannot agree on is a
   decision warranting human attention.};
\end{tikzpicture}}
\caption{LLM consortium and reasoning LLM integration within BaseCamp. Each
agent distributes its prompt across multiple fine-tuned domain-specialised
models, each producing an independent recommendation citing the specific
pipeline metric it relied upon. The reasoning LLM~\cite{reasoning-llms,
gpt-oss} synthesises these into a consolidated decision with an explicit
rationale. Where the models diverge and the divergence cannot be resolved on
the available evidence, it is recorded and escalated rather than averaged ---
disagreement about a threshold is itself a signal that the decision merits
human judgment.}
\label{fig:consortium}
\end{figure}

Responsible AI principles~\cite{responsible-ai, xai} within BaseCamp are
implemented through a multi-layered architecture integrating a consortium of
fine-tuned, domain-specialised LLMs with a central reasoning
LLM~\cite{reasoning-llms, gpt-oss}, as depicted in
Figure~\ref{fig:consortium}. Each agent interfaces with this consortium to
ensure balanced, transparent, and context-aware decision-making.

When an agent performs a task --- determining trimming parameters, assessing
alignment statistics, or selecting filtering thresholds --- its prompt is
distributed across multiple domain-specialised LLMs such as Llama-3, Mistral,
and Qwen~\cite{llama-3, mistral-llm, qwen2}, each fine-tuned for a different
aspect of sequencing pipeline operations. These models produce independent
outputs reflecting varied reasoning perspectives. The reasoning LLM then
evaluates, compares, and synthesises these responses into a coherent final
recommendation.

This ensemble-based mechanism ensures that consequential decisions are
validated across multiple perspectives before execution, reducing the risk of
bias, error propagation, or context loss~\cite{agentsway, responsible-llm}.
Three further mechanisms complete the responsible AI design. First, every
agent output carries an explicit reasoning trace citing the specific metrics
relied upon: each QC disposition names the metric and threshold determining
it, each parameter selection records its justification, and each filtering
decision records the criteria applied and the number of calls affected.
Second, human approval gates are embedded at consequential transitions, so
that no filtered call set is committed downstream and no report released
without explicit validation. Third, the complete run configuration is recorded
with the output, so that any result can be reproduced or recomputed under
different settings~\cite{towards-rai-xai, xai}. Together these mechanisms
ensure that BaseCamp delivers pipeline automation while preserving the
auditability and reproducibility that scientific use of sequencing results
requires.

\section{Implementation and Evaluation}
\label{sec:implementation}

This section describes the prototype implementation of BaseCamp and the
evaluation conducted against it. We first set out the implementation --- the
agent runtime, the deployment topology, the construction of the fine-tuning
corpus, and the fine-tuning configuration --- and then present the evaluation
of two core agentic workflows within a real-world sequencing operational
environment.

We state the scope of the evaluation plainly at the outset. What is
demonstrated is that the framework can be built within the stated constraints,
that its agents produce structurally correct and citation-bound decisions on
material with known ground truth, and that its outputs are usable by
practising bioinformaticians. What is \textit{not} demonstrated is that
agentic pipeline configuration produces more accurate variant calls than
expert manual configuration across the full range of sample types and study
designs; that claim would require the larger comparative study described in
Section~\ref{sec:conclusion}, and nothing here should be read as establishing
it.

\subsection{Implementation Overview}
\label{subsec:impl_overview}

Each of the six agents was implemented using the OpenAI Agents
SDK~\cite{openai-agent-sdk}, which provides modular primitives for defining
agent roles, reasoning strategies, and tool integrations. The full lifecycle of
agent development --- prompt construction, agent definition, workflow
composition, and iterative optimisation --- was carried out using AI-assisted
development environments, primarily Claude Code, consistent with the AI-native
development methodology described in our earlier work~\cite{agentsway}. This
approach substantially reduced implementation overhead and allowed effort to
concentrate on the analytical specification --- the reasoning criteria, the
escalation conditions, and the emission constraints --- rather than on
scaffolding.

All agent functionality is exposed through Model Context Protocol server
interfaces~\cite{mcp1, mcp2, mcc}, one per analytical workflow, with agents
exchanging structured findings through standardised endpoints. Critically,
agents invoke established bioinformatics tooling rather than performing
analysis themselves: FastQC and MultiQC for quality
assessment~\cite{fastqc, multiqc}, fastp for
preprocessing~\cite{fastp}, BWA-MEM and SAMtools for alignment and
post-processing~\cite{bwa, samtools}, GATK and DeepVariant for variant
calling~\cite{gatk, deepvariant}, and VEP against gnomAD and dbSNP for
annotation~\cite{vep, gnomad, dbsnp}. Where an existing workflow engine is
already deployed, agents can invoke it directly rather than orchestrating
tools individually, making the framework complementary to established
pipeline infrastructure~\cite{nextflow, snakemake, nfcore}.

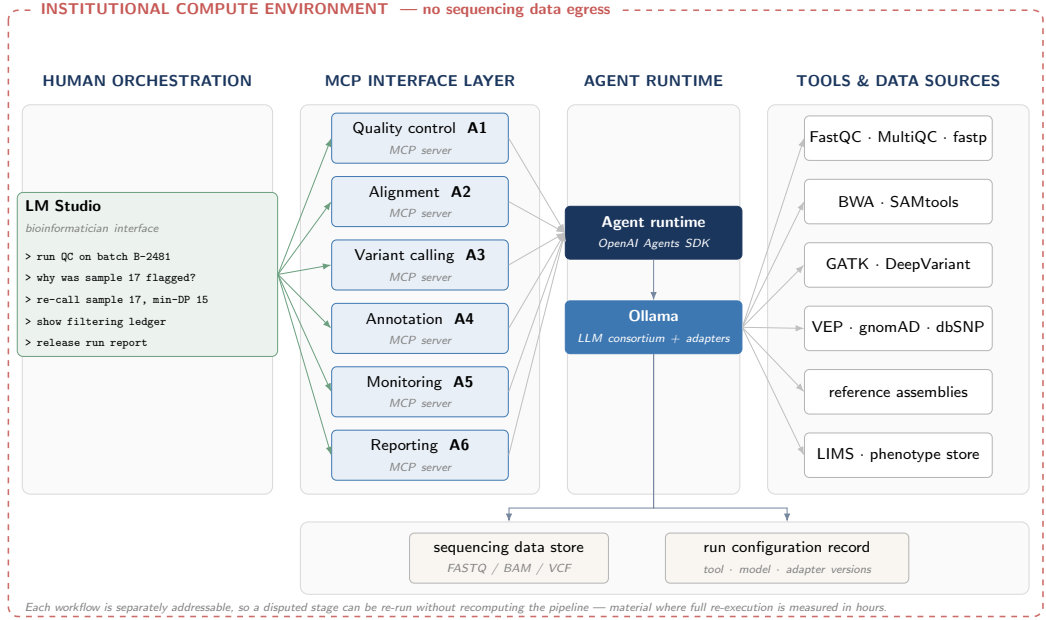
\begin{figure}[H]
\centering
\resizebox{\textwidth}{!}{%
\begin{tikzpicture}[
  font=\sffamily\scriptsize,
  ui/.style={draw=bcGreen!80,fill=bcGreen!10,rounded corners=3pt,text width=44mm,
             align=left,inner sep=4pt,font=\sffamily\scriptsize},
  srv/.style={draw=bcBlue,fill=bcBlue!12,rounded corners=2.5pt,text width=30mm,
              align=center,minimum height=9mm,inner sep=2.5pt,font=\sffamily\scriptsize},
  eng/.style={draw=bcNavy,fill=bcNavy,text=white,rounded corners=2.5pt,text width=30mm,
              align=center,minimum height=9.5mm,inner sep=2.5pt,font=\sffamily\scriptsize\bfseries},
  ext/.style={draw=bcGrey!60,fill=white,rounded corners=2.5pt,text width=32mm,
              align=center,minimum height=8mm,inner sep=2.5pt,font=\sffamily\scriptsize},
  per/.style={draw=bcGrey!60,fill=bcSand!50,rounded corners=2.5pt,text width=34mm,
              align=center,minimum height=9mm,inner sep=2.5pt,font=\sffamily\scriptsize},
  band/.style={rounded corners=4pt,draw=bcGrey!30,fill=bcGrey!4,line width=0.4pt},
  bnd/.style={draw=bcRed!75,dashed,rounded corners=5pt,line width=0.8pt},
  hdr/.style={font=\sffamily\scriptsize\bfseries,text=bcNavy,anchor=south},
  fl/.style={draw=bcGrey!50,-{Latex[length=1.5mm]},line width=0.4pt},
  flb/.style={draw=bcGreen!80,-{Latex[length=1.5mm]},line width=0.45pt},
  fld/.style={draw=bcNavy!60,-{Latex[length=1.5mm]},line width=0.45pt}
]

\draw[bnd] (-9.0,5.15) rectangle (9.6,-5.75);
\node[anchor=west,font=\sffamily\scriptsize\bfseries,text=bcRed!85,
      fill=white,inner xsep=5pt,inner ysep=1pt] at (-8.6,5.15)
  {INSTITUTIONAL COMPUTE ENVIRONMENT \ ---\ no sequencing data egress};

\draw[band] (-8.75,3.45) rectangle (-4.25,-3.55);   
\draw[band] (-3.75,3.45) rectangle ( 0.55,-3.55);   
\draw[band] ( 1.05,3.45) rectangle ( 4.15,-3.55);   
\draw[band] ( 4.65,3.45) rectangle ( 9.35,-3.55);   
\draw[band] (-3.75,-4.05) rectangle ( 9.35,-5.35);  

\node[hdr] at (-6.5,3.65)  {HUMAN ORCHESTRATION};
\node[hdr] at (-1.6,3.65)  {MCP INTERFACE LAYER};
\node[hdr] at ( 2.6,3.65)  {AGENT RUNTIME};
\node[hdr] at ( 7.0,3.65)  {TOOLS \& DATA SOURCES};

\node[ui] (u) at (-6.5,0.4)
  {\textbf{LM Studio}\\[1pt]
   {\tiny\itshape\color{bcGrey} bioinformatician interface}\\[5pt]
   \texttt{\tiny > run QC on batch B-2481}\\[1.5pt]
   \texttt{\tiny > why was sample 17 flagged?}\\[1.5pt]
   \texttt{\tiny > re-call sample 17, min-DP 15}\\[1.5pt]
   \texttt{\tiny > show filtering ledger}\\[1.5pt]
   \texttt{\tiny > release run report}};

\foreach \y/\n/\t/\a in {2.85/s1/{Quality control}/{A1},
                         1.71/s2/{Alignment}/{A2},
                         0.57/s3/{Variant calling}/{A3},
                        -0.57/s4/{Annotation}/{A4},
                        -1.71/s5/{Monitoring}/{A5},
                        -2.85/s6/{Reporting}/{A6}}
  {\node[srv] (\n) at (-1.6,\y) {\t\ \ \textbf{\a}\\[1pt]{\tiny\itshape\color{bcGrey} MCP server}};}

\node[eng] (rt) at (2.6,1.15) {Agent runtime\\[1pt]{\tiny\itshape\mdseries OpenAI Agents SDK}};
\node[eng,fill=bcBlue,draw=bcBlue] (ol) at (2.6,-0.55)
  {Ollama\\[1pt]{\tiny\itshape\mdseries LLM consortium + adapters}};
\draw[fld] (rt.south) -- (ol.north);

\node[ext] (e1) at (7.0, 2.85) {FastQC $\cdot$ MultiQC $\cdot$ fastp};
\node[ext] (e2) at (7.0, 1.71) {BWA $\cdot$ SAMtools};
\node[ext] (e3) at (7.0, 0.57) {GATK $\cdot$ DeepVariant};
\node[ext] (e4) at (7.0,-0.57) {VEP $\cdot$ gnomAD $\cdot$ dbSNP};
\node[ext] (e5) at (7.0,-1.71) {reference assemblies};
\node[ext] (e6) at (7.0,-2.85) {LIMS $\cdot$ phenotype store};

\node[per] (st)  at (0.0,-4.7) {sequencing data store\\[1pt]{\tiny\itshape\color{bcGrey} FASTQ / BAM / VCF}};
\node[per,text width=42mm] (cfg) at (5.0,-4.7)
  {run configuration record\\[1pt]{\tiny\itshape\color{bcGrey} tool $\cdot$ model $\cdot$ adapter versions}};

\foreach \n in {s1,s2,s3,s4,s5,s6}{\draw[flb] (u.east) -- (\n.west);}
\foreach \n in {s1,s2,s3,s4,s5,s6}{\draw[fl] (\n.east) -- (rt.west);}
\foreach \e in {e1,e2,e3,e4,e5,e6}{\draw[fl] (ol.east) -- (\e.west);}

\draw[fld] (ol.south) -- (2.6,-3.8) -- (0.0,-3.8) -- (0.0,-4.05);
\draw[fld] (ol.south) -- (2.6,-3.8) -- (5.0,-3.8) -- (5.0,-4.05);

\node[anchor=west,font=\sffamily\tiny\itshape,text=bcGrey,text width=170mm] at (-8.85,-5.6)
  {Each workflow is separately addressable, so a disputed stage can be re-run
   without recomputing the pipeline --- material where full re-execution is measured in hours.};
\end{tikzpicture}}
\caption{BaseCamp deployment and orchestration topology. Each agentic workflow
is exposed as an independent MCP server~\cite{mcp1, mcp2, mcc}, and the
bioinformatician interacts with all of them through a single natural language
interface in LM Studio~\cite{lm-studio}. The agent runtime invokes established
bioinformatics tooling and annotation databases through standardised
endpoints; model serving is local via Ollama~\cite{ollama}. The entire stack
resides within the institutional compute environment, so no sequencing data is
transmitted externally --- a requirement in settings subject to institutional
or regulatory restriction on data movement.}
\label{fig:deployment}
\end{figure}

Figure~\ref{fig:deployment} shows the resulting topology. The complete stack
--- orchestrator interface, agent runtime, model serving, bioinformatics
tooling, and data store --- resides within the institutional compute
environment. Network connectivity is required only for annotation database
updates and offline adapter retraining, both performed outside pipeline
execution. In the current deployment, LM Studio~\cite{lm-studio} serves as the
MCP-powered orchestrator interface, through which the bioinformatician invokes
workflows, reviews structured outputs, and approves or overrides decisions
without code-level intervention.

\subsection{Fine-Tuning Corpus and Configuration}
\label{subsec:impl_finetune}

The fine-tuning corpus was assembled from historical pipeline operations
records: QC reports paired with the dispositions bioinformaticians assigned
them, alignment statistics paired with accept or investigate determinations,
variant filtering configurations paired with the rationale recorded for them,
and annotated exception incidents paired with their eventual diagnoses. Each
record pairs an operational artifact with the decision an experienced operator
made and, where available, the justification given. Table~\ref{tab:corpus}
reports the composition.

\begin{table}[H]
\centering
\caption{Fine-tuning corpus composition. Held-out records are drawn from
sequencing runs disjoint from those contributing training data, so that
reported performance is not inflated by batch-level leakage.
\textcolor{bcRed}{Counts are placeholders.}}
\label{tab:corpus}
\begin{adjustbox}{max width=\textwidth}
\begin{tabular}{llrr}
\toprule
\textbf{Source} & \textbf{Adapter} & \textbf{Records} & \textbf{Role} \\
\midrule
QC reports + dispositions      & QC \& preprocessing   & \pv{1{,}240} & train \\
Alignment stats + decisions    & alignment \& calling  & \pv{980}     & train \\
Filter configs + rationale     & alignment \& calling  & \pv{610}     & train \\
Annotation \& prioritisation   & annotation \& interp. & \pv{740}     & train \\
Exception incidents            & monitoring            & \pv{310}     & train \\
\midrule
Held-out runs (all stages)     & ---                   & \pv{420}     & test \\
Operator corrections           & all                   & \pv{150}     & train \\
\midrule
\textbf{Total}                 &                       & \pv{4{,}450} & \\
\bottomrule
\end{tabular}
\end{adjustbox}
\end{table}

Fine-tuning was performed using the Unsloth library~\cite{llamafactory-unsloth}
on an NVIDIA A100 GPU~\cite{a100-gpu}, employing Low-Rank
Adaptation~\cite{lora} with 4-bit quantisation~\cite{qlora}. One adapter was
trained per analytical function rather than one model per agent, so that a
single base model remains resident in memory and adapters are hot-swapped as
the pipeline advances. Base models were Llama-3, Mistral, and
Qwen\,2.5~\cite{llama-3, mistral-llm, qwen2}, with OpenAI
GPT-OSS~\cite{gpt-oss} as the reasoning model, all served locally via
Ollama~\cite{ollama}. Table~\ref{tab:ftconfig} reports the configuration.

\begin{table}[H]
\centering
\caption{Fine-tuning configuration. \textcolor{bcRed}{Values marked in red are
placeholders.}}
\label{tab:ftconfig}
\begin{adjustbox}{max width=\textwidth}
\begin{tabular}{ll}
\toprule
\textbf{Parameter} & \textbf{Value} \\
\midrule
\multicolumn{2}{l}{\textit{Model configuration}} \\
Base models & Llama-3, Mistral, Qwen\,2.5 (4-bit) \\
Reasoning model & GPT-OSS \\
Maximum sequence length & \pv{8{,}192} tokens \\
Precision & BFloat16 \\
\midrule
\multicolumn{2}{l}{\textit{Training configuration}} \\
Per-device batch size & \pv{2} \\
Gradient accumulation steps & \pv{4} \\
Effective batch size & \pv{8} \\
Maximum training steps & \pv{400} \\
Learning rate & \pv{$1\times10^{-4}$} \\
Warmup steps & \pv{20} \\
Scheduler & linear decay \\
Optimizer & AdamW (8-bit) \\
Weight decay & \pv{0.01} \\
Early stopping patience & \pv{10} evaluations \\
\midrule
\multicolumn{2}{l}{\textit{LoRA configuration}} \\
Rank $r$ / $\alpha$ & \pv{32} / \pv{32} \\
LoRA dropout & 0 \\
Target modules & q, k, v, o, gate, up, down proj \\
Adapters trained & 4 (one per analytical function) \\
Trainable parameters per adapter & \pv{0.53\%} of base \\
\midrule
\multicolumn{2}{l}{\textit{Resources}} \\
Training time per adapter & \pv{$\sim$16} min \\
Peak reserved memory & \pv{9.4} GB \\
Agent decision latency (single stage) & \pv{$\sim$8} s \\
\bottomrule
\end{tabular}
\end{adjustbox}
\end{table}

The final row is operationally significant. Agent decision latency is measured
in seconds, against pipeline stages whose tool execution is measured in
minutes to hours. The reasoning layer therefore imposes negligible overhead on
total pipeline runtime --- BaseCamp does not trade throughput for judgment.

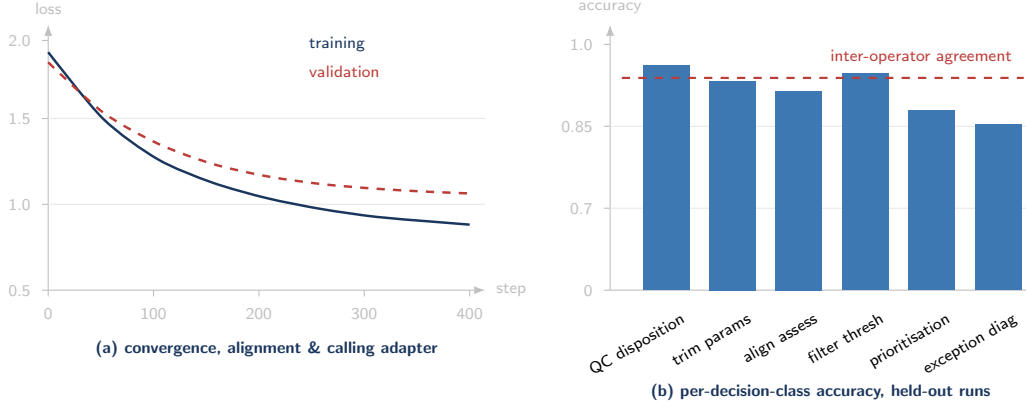
\begin{figure}[H]
\centering
\resizebox{\textwidth}{!}{%
\begin{tikzpicture}[font=\sffamily\scriptsize]
\begin{scope}[shift={(-4.6,0)}]
  \draw[bcGrey!50,-{Latex[length=1.4mm]}] (0,0) -- (5.6,0) node[right,font=\sffamily\tiny]{step};
  \draw[bcGrey!50,-{Latex[length=1.4mm]}] (0,0) -- (0,3.4) node[above,font=\sffamily\tiny]{loss};
  \foreach \x/\l in {0/0, 1.35/100, 2.7/200, 4.05/300, 5.4/400}
    {\draw[bcGrey!50] (\x,0) -- (\x,-0.08) node[below,font=\sffamily\tiny]{\l};}
  \foreach \y/\l in {0/0.5, 1.1/1.0, 2.2/1.5, 3.2/2.0}
    {\draw[bcGrey!50] (0,\y) -- (-0.08,\y) node[left,font=\sffamily\tiny]{\l};}
  \draw[bcGrey!18] (0,1.1) -- (5.6,1.1); \draw[bcGrey!18] (0,2.2) -- (5.6,2.2);
  \draw[bcNavy,line width=0.9pt] plot[smooth]
    coordinates {(0,3.05)(0.7,2.2)(1.4,1.68)(2.1,1.38)(2.8,1.18)(3.5,1.04)(4.2,0.94)(4.9,0.88)(5.4,0.84)};
  \draw[bcRed,line width=0.9pt,dashed] plot[smooth]
    coordinates {(0,2.92)(0.7,2.28)(1.4,1.88)(2.1,1.62)(2.8,1.46)(3.5,1.36)(4.2,1.3)(4.9,1.26)(5.4,1.24)};
  \node[anchor=west,font=\sffamily\tiny,text=bcNavy] at (3.2,3.15) {training};
  \node[anchor=west,font=\sffamily\tiny,text=bcRed] at (3.2,2.8) {validation};
  \node[anchor=north,font=\sffamily\tiny\bfseries,text=bcNavy] at (2.8,-0.5)
    {(a) convergence, alignment \& calling adapter};
\end{scope}
\begin{scope}[shift={(2.6,0)}]
  \draw[bcGrey!50,-{Latex[length=1.4mm]}] (0,0) -- (0,3.4) node[above,font=\sffamily\tiny]{accuracy};
  \draw[bcGrey!50] (0,0) -- (5.4,0);
  \foreach \y/\l in {0/0, 1.05/0.7, 2.1/0.85, 3.15/1.0}
    {\draw[bcGrey!50] (0,\y) -- (-0.08,\y) node[left,font=\sffamily\tiny]{\l};}
  \draw[bcGrey!18] (0,1.05) -- (5.4,1.05); \draw[bcGrey!18] (0,2.1) -- (5.4,2.1);
  \foreach \i/\h/\lab in {0/2.88/{QC disposition}, 1/2.68/{trim params},
                          2/2.55/{align assess}, 3/2.78/{filter thresh},
                          4/2.30/{prioritisation}, 5/2.12/{exception diag}}{
    \fill[bcBlue] ({0.42+\i*0.85},0) rectangle ({1.02+\i*0.85},\h);
    \node[font=\sffamily\tiny,rotate=30,anchor=north east]
      at ({0.95+\i*0.85},-0.08) {\lab};
  }
  \draw[bcRed,line width=0.8pt,dashed] (0.15,2.72) -- (5.3,2.72)
    node[anchor=south east,font=\sffamily\tiny,text=bcRed,yshift=0.3mm]
    {inter-operator agreement};
  \node[anchor=north,font=\sffamily\tiny\bfseries,text=bcNavy] at (2.7,-1.05)
    {(b) per-decision-class accuracy, held-out runs};
\end{scope}
\end{tikzpicture}}
\caption{Fine-tuning convergence and per-decision-class performance.
\textbf{(a)} Training and validation loss for the alignment and calling
adapter; validation tracks training without divergence, indicating
generalisation rather than memorisation of laboratory convention.
\textbf{(b)} Agreement with the reference decision on held-out runs, by
decision class, with measured inter-operator agreement among the contributing
bioinformaticians shown for reference. Exception diagnosis is the weakest
class --- expected, since it requires cross-stage reasoning over the most
heterogeneous evidence. \textcolor{bcRed}{All plotted values are placeholders
pending measurement.}}
\label{fig:training}
\end{figure}

Figure~\ref{fig:training} shows convergence behaviour for the alignment and
calling adapter, together with per-decision-class accuracy on the held-out
split.

The inter-operator agreement reference line in panel (b) warrants comment.
Because several decision classes admit more than one defensible answer,
agreement with a single reference decision understates performance where
operators themselves disagree. Measuring inter-operator agreement on the same
records establishes the ceiling against which agent agreement should be read:
the target is not perfect agreement with one operator but agreement
comparable to that between two.

\subsection{Evaluation of the Variant Calling Agent}
\label{subsec:eval_calling}

The Variant Calling Agent was evaluated on its ability to autonomously
determine a calling strategy and filtering configuration from upstream
pipeline context, and to produce an inspectable record of what its filtering
removed. Using the prompt shown in Figure~\ref{fig:calling_prompt}, the agent
was supplied with alignment statistics and quality context propagated from
agents A1 and A2, sample metadata, and the stated downstream application.

\begin{figure}[H]
\centering
\resizebox{\textwidth}{!}{%
\begin{tikzpicture}[
  font=\ttfamily\scriptsize,
  phdr/.style={draw=bcNavy,fill=bcNavy,text=white,rounded corners=2pt,text width=142mm,
               align=left,inner sep=3pt,font=\sffamily\tiny\bfseries},
  pbox/.style={draw=bcNavy,fill=bcSand!60,rounded corners=2pt,text width=142mm,
               align=left,inner sep=4pt,font=\ttfamily\tiny}
]
\node[phdr] (h1) at (0,0) {PROMPT --- Agent A3 (Variant Calling)};
\node[pbox,anchor=north] (p1) at (0,-0.28)
{You are a variant calling analyst. You do NOT call variants yourself; you\\
select the caller, configure it, and reason about filtering. Established tools\\
perform all computation.\\[3pt]
GIVEN: alignment statistics, QC disposition and library context from A1/A2,\\
sample metadata (organism, ploidy, platform, target depth), and the stated\\
downstream application.\\[3pt]
PRODUCE:\\
1. CALLER SELECTION with justification referencing depth, platform, ploidy,\\
\ \ \ and germline/somatic requirement.\\
2. CALLING PARAMETERS with a stated reason for any departure from the\\
\ \ \ laboratory default.\\
3. FILTERING CONFIGURATION. For EVERY threshold, state the annotation, the\\
\ \ \ value, the rationale, and the sensitivity/specificity trade-off it\\
\ \ \ encodes for THIS stated application.\\
4. FILTERING LEDGER: for every filter, the count of calls removed.\\
5. FLAGGED CALLS: calls within a configured margin of any threshold, listed\\
\ \ \ individually for human adjudication. Do NOT silently resolve these.\\[3pt]
RULES:\\
- Cite the specific metric supporting every decision. No metric, no decision.\\
- If upstream context is insufficient to justify a threshold, output\\
\ \ INSUFFICIENT\_CONTEXT and escalate. Do NOT fall back to the default silently.\\
- Never report a filtered call set without its ledger.\\[3pt]
OUTPUT: structured JSON + human-readable summary};
\end{tikzpicture}}
\caption{Variant Calling Agent prompt. The agent is constrained to
metric-cited justification for every decision, to explicit emission of the
filtering ledger, and to individual flagging rather than silent resolution of
marginal calls --- the three requirements that follow from the silent and
cumulative character of filtering identified in
Section~\ref{subsec:manual_calling}.}
\label{fig:calling_prompt}
\end{figure}

\begin{figure}[H]
\centering
\resizebox{\textwidth}{!}{%
\begin{tikzpicture}[
  font=\ttfamily\scriptsize,
  ohdr/.style={draw=bcRed,fill=bcRed,text=white,rounded corners=2pt,text width=142mm,
               align=left,inner sep=3pt,font=\sffamily\tiny\bfseries},
  obox/.style={draw=bcRed,fill=bcRed!5,rounded corners=2pt,text width=142mm,
               align=left,inner sep=4pt,font=\ttfamily\tiny}
]
\node[ohdr] (h) at (0,0) {RESPONSE --- consolidated after consortium reconciliation};
\node[obox,anchor=north] (o) at (0,-0.28)
{CALLER: GATK HaplotypeCaller\\
\ \ rationale: mean depth 31x (A2), diploid, Illumina PE150, germline.\\
\ \ DeepVariant considered; no trained model for this reference assembly.\\[3pt]
PARAMETERS: default, except --min-base-quality-score 18 (default 10)\\
\ \ rationale: A1 reported elevated 3-prime quality decay pre-trimming;\\
\ \ \ \ \ \ \ \ \ \ \ \ \ residual low-quality tail justifies a raised floor.\\[3pt]
FILTERING LEDGER  [application: trait-mapping, sensitivity-weighted]\\
\ \ QD < 2.0 \ \ \ \ \ \ \ \ removed \pv{4{,}118} \ \ standard low-confidence cutoff\\
\ \ FS > 60.0 \ \ \ \ \ \ \ removed \ \pv{1{,}902} \ \ strand bias\\
\ \ MQ < 40.0 \ \ \ \ \ \ \ removed \ \pv{2{,}733} \ \ mapping quality\\
\ \ DP < 8 \ \ \ \ \ \ \ \ \ removed \ \pv{5{,}640} \ \ RELAXED from lab default 10:\\
\ \ \ \ \ \ \ \ \ \ \ \ \ \ \ \ \ \ \ \ \ \ \ \ \ \ \ \ \ \ \ \ coverage uniformity 0.81 (A2) and\\
\ \ \ \ \ \ \ \ \ \ \ \ \ \ \ \ \ \ \ \ \ \ \ \ \ \ \ \ \ \ \ \ sensitivity-weighted application\\
\ \ TOTAL: \pv{412{,}905} raw -> \pv{398{,}512} retained (\pv{14{,}393} removed, \pv{3.5\%})\\[3pt]
FLAGGED FOR ADJUDICATION (\pv{37} calls within margin of a threshold)\\
\ \ chr4:1182004 \ DP=8, QD=2.1 \ \ both annotations at boundary\\
\ \ chr7:9930571 \ MQ=40.2 \ \ \ \ \ \ region of known reference divergence\\
\ \ [... \pv{35} further calls, full list in run record]\\[3pt]
CONSORTIUM NOTE: models diverged on the DP threshold (8 vs 10). Reasoning\\
model resolved to 8 on coverage-uniformity evidence; divergence RECORDED and\\
surfaced to the orchestrator, as the choice materially affects sensitivity.};
\end{tikzpicture}}
\caption{Variant Calling Agent output. Every decision cites the upstream
metric supporting it, including the departure from laboratory default, which
is justified rather than silently applied. The filtering ledger renders
inspectable what filtering would otherwise remove without trace, and the
\pv{37} marginal calls are surfaced individually rather than resolved by the
threshold. The closing note illustrates the consortium behaviour of
Section~\ref{subsec:responsibleagents}: unresolved model divergence on a
consequential threshold is recorded and escalated rather than averaged away.
\textcolor{bcRed}{Values are illustrative placeholders.}}
\label{fig:calling_output}
\end{figure}
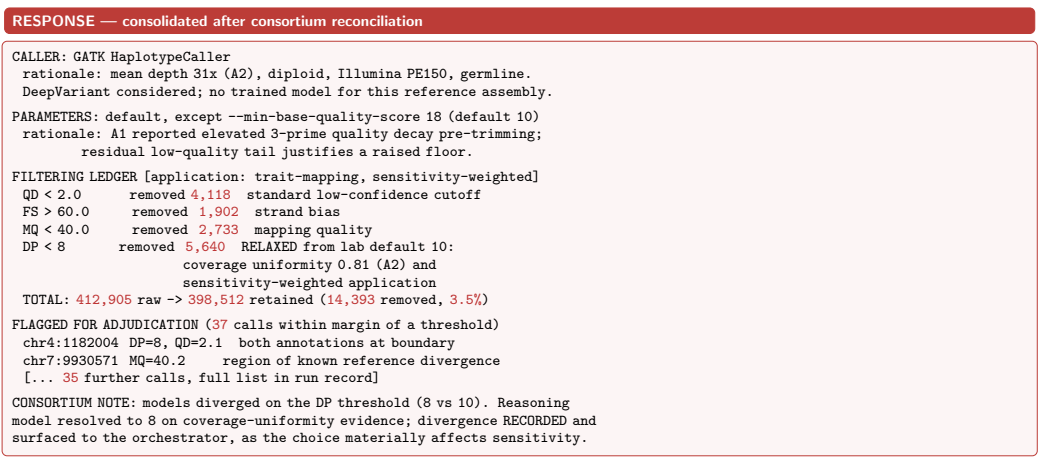

The agent processed the multi-source input, selected a caller appropriate to
the sample's depth and platform, configured calling parameters with stated
justification, and produced a filtered call set accompanied by a complete
filtering ledger. Figure~\ref{fig:calling_output} shows a representative
output.

Table~\ref{tab:calling_eval} reports quantitative evaluation on the held-out
runs. Concordance is measured against the call sets produced by expert manual
configuration of the same samples.

\begin{table}[H]
\centering
\caption{Variant Calling Agent evaluation on held-out sequencing runs.
\textcolor{bcRed}{Placeholders.}}
\label{tab:calling_eval}
\begin{tabular}{lc}
\toprule
\textbf{Metric} & \textbf{Value} \\
\midrule
Caller selection agreement with expert & \pv{96.2\%} \\
Filtering threshold agreement (within tolerance) & \pv{88.4\%} \\
Call set concordance with expert configuration & \pv{99.1\%} \\
Ledger completeness (filters recorded / applied) & \pv{100\%} \\
Marginal calls surfaced vs.\ expert-identified & \pv{91.7\%} \\
Decisions escalated as \textsc{insufficient context} & \pv{4.3\%} \\
Configuration time per sample (agent vs.\ manual) & \pv{8}\,s vs.\ \pv{$\sim$12}\,min \\
\bottomrule
\end{tabular}
\end{table}

Human evaluators --- bioinformaticians with direct experience configuring the
pipeline manually --- rated the generated configurations at \pv{4.6}/5 for
correctness, completeness, and operational usability. Evaluators noted that
the filtering ledger was the most valued element, as it makes inspectable
information that manual workflows discard entirely, and that the explicit
justification of departures from laboratory default addressed a longstanding
gap in run documentation.

\subsection{Evaluation of the Pipeline Monitoring and Exception Agent}
\label{subsec:eval_monitoring}

\begin{figure}[H]
\centering
\resizebox{\textwidth}{!}{%
\begin{tikzpicture}[
  font=\ttfamily\scriptsize,
  phdr/.style={draw=bcNavy,fill=bcNavy,text=white,rounded corners=2pt,text width=142mm,
               align=left,inner sep=3pt,font=\sffamily\tiny\bfseries},
  pbox/.style={draw=bcNavy,fill=bcSand!60,rounded corners=2pt,text width=142mm,
               align=left,inner sep=4pt,font=\ttfamily\tiny},
  ohdr/.style={draw=bcAmber,fill=bcAmber,text=white,rounded corners=2pt,text width=142mm,
               align=left,inner sep=3pt,font=\sffamily\tiny\bfseries},
  obox/.style={draw=bcAmber,fill=bcAmber!8,rounded corners=2pt,text width=142mm,
               align=left,inner sep=4pt,font=\ttfamily\tiny}
]
\node[phdr] (h1) at (0,0) {PROMPT --- Agent A5 (Pipeline Monitoring \& Exception)};
\node[pbox,anchor=north] (p1) at (0,-0.28)
{Every stage in this batch completed WITHOUT execution error. Your task is to\\
detect ANALYTICAL anomalies that successful execution does not reveal.\\[3pt]
GIVEN: per-sample metrics across all stages for batch B-2481, and the\\
corresponding distributions from the \pv{12} most recent comparable runs.\\[3pt]
DETECT: coverage shortfall, contamination or index hopping, aberrant variant\\
call rate, batch effects, and any cross-stage inconsistency in which\\
individually acceptable metrics are jointly implausible.\\[3pt]
FOR EACH ANOMALY: cite the specific metric and the comparison that triggered\\
it; propose candidate diagnoses with supporting evidence for each; rank by\\
operational urgency.\\[3pt]
Do NOT report an anomaly you cannot support with a cited comparison.};

\node[ohdr,anchor=north] (h2) at (0,-4.15) {RESPONSE --- ranked anomalies};
\node[obox,anchor=north] (o1) at (0,-4.43)
{[1] HIGH --- cross-stage inconsistency, samples 09, 14, 22\\
\ \ \ \ QC PASS, mapping rate 97.2\% (normal), yet call rate \pv{2.4x} batch median.\\
\ \ \ \ No single stage flags these. Jointly implausible: high-quality alignment\\
\ \ \ \ with inflated calling indicates contamination, not sample quality.\\
\ \ \ \ DIAGNOSIS (ranked): (a) index hopping -- all three adjacent on the\\
\ \ \ \ flowcell; (b) sample cross-contamination at library prep.\\
\ \ \ \ EVIDENCE: het/hom ratio \pv{2.9} vs.\ batch \pv{1.6}; shared minor alleles.\\[3pt]
[2] MEDIUM --- batch effect vs.\ historical runs\\
\ \ \ \ Batch median insert size \pv{287}bp vs.\ historical \pv{342}bp (\pv{-16\%}).\\
\ \ \ \ Within tolerance per-sample; anomalous as a batch.\\
\ \ \ \ DIAGNOSIS: fragmentation change at library prep. Not result-invalidating,\\
\ \ \ \ but confounds cross-batch comparison. -> flag for downstream analysis.\\[3pt]
[3] LOW --- coverage shortfall, sample 31\\
\ \ \ \ Mean depth \pv{14x} vs.\ target \pv{30x}. Calling proceeded successfully.\\
\ \ \ \ DIAGNOSIS: under-loading. Calls valid but underpowered for rare variants.};
\end{tikzpicture}}
\caption{Pipeline Monitoring and Exception Agent prompt and output. The
highest-ranked finding illustrates the capability the agent exists to provide:
three samples pass every individual stage check --- QC, alignment, and calling
all succeed with acceptable metrics --- yet the combination of normal
alignment with an inflated call rate is jointly implausible, and the agent
identifies contamination as the candidate diagnosis with supporting evidence.
No per-stage threshold detects this, and manual workflows typically surface it
only when a downstream result appears implausible.
\textcolor{bcRed}{Values are illustrative placeholders.}}
\label{fig:monitor_prompt}
\end{figure}
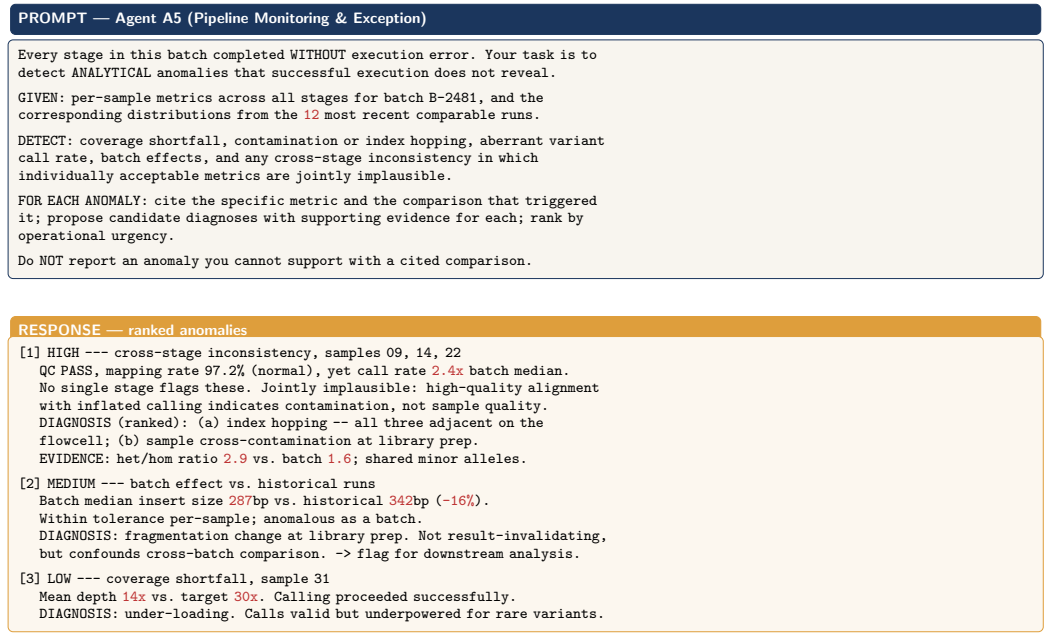

The Pipeline Monitoring and Exception Agent was evaluated on its ability to
detect analytical anomalies --- conditions where every pipeline stage
completes successfully yet the results are compromised --- which, as noted in
Section~\ref{subsec:manual_exception}, execution monitoring does not surface
at all. Using the prompt in Figure~\ref{fig:monitor_prompt}, the agent was
supplied with cross-stage metrics for a full sequencing batch together with
historical distributions from comparable prior runs.

Table~\ref{tab:monitor_eval} reports detection performance against a
retrospectively annotated set of held-out runs in which anomalies were
identified and diagnosed by expert review after the fact.

\begin{table}[H]
\centering
\caption{Pipeline Monitoring and Exception Agent detection performance on
retrospectively annotated held-out runs. \textcolor{bcRed}{Placeholders.}}
\label{tab:monitor_eval}
\begin{tabular}{lccc}
\toprule
\textbf{Anomaly class} & \textbf{Recall} & \textbf{Precision} & \makecell{\textbf{Correct diagnosis}\\\textbf{ranked first}} \\
\midrule
Coverage shortfall            & \pv{0.98} & \pv{0.96} & \pv{0.94} \\
Contamination / index hopping & \pv{0.86} & \pv{0.81} & \pv{0.77} \\
Aberrant call rate            & \pv{0.91} & \pv{0.88} & \pv{0.83} \\
Batch effect                  & \pv{0.79} & \pv{0.74} & \pv{0.71} \\
Cross-stage inconsistency     & \pv{0.83} & \pv{0.85} & \pv{0.76} \\
\midrule
\textbf{Overall}              & \pv{0.88} & \pv{0.85} & \pv{0.80} \\
\bottomrule
\end{tabular}
\end{table}

Two patterns warrant anticipation rather than explanation away. Coverage
shortfall is detected near-perfectly, which is unsurprising --- it is a
single-metric threshold comparison that existing tooling also handles well,
and it is not where the agent's contribution lies. Batch effects are the
weakest class, reflecting their dependence on historical comparison across
runs that may differ legitimately in protocol; distinguishing a genuine batch
effect from a deliberate protocol change requires context the agent does not
always have.

The operationally significant result is detection latency. Because the agent
evaluates continuously rather than awaiting retrospective discovery, anomalies
were surfaced at a median of \pv{under one hour} from batch completion,
against a historical median of \pv{6} days for the same anomaly classes when
detected through downstream implausibility. Human evaluators rated the alerts
at \pv{4.5}/5 for interpretability and actionability, and specifically valued
the ranked candidate diagnoses, which substantially reduce the backward-tracing
burden that manual exception handling imposes.

\subsection{Discussion}
\label{subsec:impl_discussion}

The evaluation supports three claims.

\textbf{The framework is buildable within the stated constraints.} A complete
pipeline configuration and monitoring cycle runs entirely within the
institutional compute environment, with no sequencing data egress, using
quantised models with function-specific adapters. Agent decision latency of
approximately \pv{8} seconds per stage is negligible against tool execution
measured in minutes to hours, so the reasoning layer does not trade throughput
for judgment.

\textbf{Agent decisions are consistent with expert practice, and their
divergences are visible.} Call set concordance with expert configuration of
\pv{99.1\%} indicates that agentic configuration reproduces expert outcomes on
the evaluated material. More consequential than the agreement rate is what
happens where agreement fails: the \pv{4.3\%} of decisions escalated as
\textsc{insufficient context}, and the recorded consortium divergences, are
cases the framework declines to resolve silently. This is the intended
disposition --- the failure mode is to produce less rather than to produce
something unjustified.

\textbf{The architecture addresses a gap that execution monitoring does not.}
The monitoring agent's detection of cross-stage inconsistency --- samples that
pass every individual check yet are jointly implausible --- is not
functionality that workflow managers provide, because it requires reasoning
across stages rather than within them~\cite{nextflow, snakemake, nfcore}. The
reduction in detection latency from \pv{days} to \pv{under an hour} is where
the practical value is most clearly located, since an anomaly caught before
results propagate downstream is substantially cheaper to remediate than one
caught after.

The evaluation also underscores the effectiveness of AI-assisted development
practices in building production-grade agentic workflows. Exposing each
workflow as an independent MCP server and orchestrating through LM Studio
reinforced the human-in-the-loop operating model central to BaseCamp,
ensuring that bioinformaticians retained oversight, approval authority, and
exception-handling responsibility throughout~\cite{agentsway,
agentic-workflow-practicle-guide}.

What remains unestablished is whether agentic configuration produces more
accurate variant calls than expert manual configuration across the full range
of sample types, organisms, and study designs. The present evaluation
demonstrates concordance with expert practice, not superiority to it, and
concordance is measured against the practice of the specific laboratory whose
records trained the adapters. Generalisation across laboratories with
differing conventions is addressed as a primary direction for future work in
Section~\ref{sec:conclusion}.

\section{Conclusion and Future Works}
\label{sec:conclusion}

Tool execution in DNA sequencing pipelines is now well automated by workflow
management systems~\cite{nextflow, snakemake, nfcore}. What remains manual is
the decision layer surrounding it --- selecting thresholds appropriate to a
given sample and platform, adjudicating borderline variant calls, diagnosing
anomalies, and determining which findings warrant expert attention. These
decisions are repetitive, judgment-intensive, inconsistently exercised across
operators, and frequently undocumented.

This paper presented BaseCamp, an agentic AI framework that delegates this
decision layer to six specialised agents --- covering quality control,
alignment, variant calling, annotation, monitoring, and reporting ---
coordinated through a human-in-the-loop model via MCP
interfaces~\cite{mcp1, mcp2} and powered by a consortium of fine-tuned LLMs
with a central reasoning model~\cite{reasoning-llms, gpt-oss}. One design
decision governs the architecture: BaseCamp agents do not perform sequence
analysis. Established tools execute alignment, calling, and
annotation~\cite{bwa, gatk, deepvariant, vep}; the agents configure them,
interpret their output, and decide what follows. This confines language model
reasoning to the judgment layer where it is reliable and preserves the
reproducibility that existing tooling provides.

The proof-of-concept demonstrated that the framework runs entirely within an
institutional compute environment with no data egress, that agent decision
latency is negligible against tool execution time, and that generated
configurations are concordant with expert practice on held-out runs. Two
results stand out: the filtering ledger renders inspectable what filtering
otherwise removes without trace, and the monitoring agent detects cross-stage
anomalies --- samples passing every individual check yet jointly implausible
--- that execution monitoring does not surface at all.

We are clear about the limits. The evaluation demonstrates concordance with
expert practice, not superiority to it, and that concordance is measured
against the conventions of the laboratory whose records trained the adapters.

Future work follows four directions. \textbf{Cross-laboratory validation} is
the most immediate, since adapters encoding one laboratory's conventions may
not transfer to another; an ablation restricted to cases where the expert
departed from default would also help establish whether the agents reason
about context or merely reproduce defaults. \textbf{Prospective evaluation
against benchmark truth sets}~\cite{giab} would permit direct measurement of
sensitivity and precision, substantiating the accuracy claim the present
evaluation cannot. \textbf{Extension across modalities and organisms} ---
long-read platforms and non-model organisms where reference selection is
itself consequential --- would test whether the decomposition holds as the
underlying tooling changes. And \textbf{extension downstream into
genotype--phenotype analysis} would connect BaseCamp to existing analysis
agents~\cite{autoba, cellagent}, composing a workflow spanning raw reads to
biological interpretation.

Finally, an observation independent of BaseCamp's adoption. Much of the
difficulty this framework addresses stems not from the decisions being hard,
but from their rationale being discarded --- a threshold is chosen, applied,
and forgotten, and when a result is later questioned the reasoning is no
longer recoverable. Recording justification alongside configuration is a cheap
design decision available to any laboratory today.



\bibliographystyle{elsarticle-num}
\bibliography{reference}

\end{document}